%% file: neurips_2025.tex
\documentclass{article}

\PassOptionsToPackage{square, comma, numbers,sort}{natbib}

\usepackage[preprint]{neurips_2025}

\usepackage[utf8]{inputenc} 
\usepackage[T1]{fontenc}    
\usepackage{url}            
\usepackage{booktabs}       
\usepackage{amsfonts}       
\usepackage{nicefrac}       
\usepackage{microtype}      
\usepackage{graphicx}
\usepackage{amsmath}
\usepackage{subcaption}
\usepackage{color}
\usepackage{multirow}
\usepackage{enumitem}
\usepackage[table,dvipsnames]{xcolor}
\definecolor{lightorange}{RGB}{255, 216, 177}
\usepackage{authblk}

\usepackage{xspace}
\newcommand{\suite}{INT-ACT\xspace}

\usepackage[pagebackref=true,breaklinks=true,colorlinks,bookmarks=false]{hyperref}
\hypersetup{
linkcolor=BrickRed
,citecolor=Green
,filecolor=Mulberry
,urlcolor=NavyBlue
,menucolor=BrickRed
,runcolor=Mulberry
,linkbordercolor=BrickRed
,citebordercolor=Green
,filebordercolor=Mulberry
,urlbordercolor=NavyBlue
,menubordercolor=BrickRed
,runbordercolor=Mulberry
}
\usepackage[most]{tcolorbox}  
\usepackage{xcolor}           
\usepackage{parskip}          

\newcommand{\question}[2]{
    \vspace{-0.1cm}
    \begin{tcolorbox}[
        colback=white!90!gray,      
        colframe=teal!60!black,     
        arc=5pt,                    
        boxsep=5pt,                 
        left=10pt,                  
        right=10pt,                 
        top=2pt,                    
        bottom=2pt,                 
        boxrule=0.8pt,              
        drop shadow=gray!50!white,  
        enhanced jigsaw             
    ]
    \vspace{-0.1cm}
        \paragraph{\textbf{\textit{Question #1:}}} #2
    \vspace{-0.1cm}
    \end{tcolorbox}
    \vspace{-0.1cm}
}

\title{From Intention to Execution: \\ Probing the Generalization Boundaries of Vision-Language-Action Models}

\author{
  \textbf{Irving Fang}\thanks{Equal contribution} \quad
  \textbf{Juexiao Zhang}$^{\ast}$\thanks{Project Lead} \quad
  \textbf{Shengbang Tong} \quad
  \textbf{Chen Feng} \\
  New York University \\
}

\begin{document}

\maketitle

\begin{abstract}
One promise that Vision-Language-Action (VLA) models hold over traditional imitation learning for robotics is to leverage the broad generalization capabilities of large Vision-Language Models (VLMs) to produce versatile, “generalist” robot policies. However, current evaluations of VLAs remain insufficient. Traditional imitation learning benchmarks are unsuitable due to the lack of language instructions. Emerging benchmarks for VLAs that incorporate language often come with limited evaluation tasks and do not intend to investigate how much VLM pretraining truly contributes to the generalization capabilities of the downstream robotic policy. Meanwhile, much research relies on real-world robot setups designed in isolation by different institutions, which creates a barrier for reproducibility and accessibility. To address this gap, we introduce a unified probing suite of 50 simulation‐based tasks across 10 subcategories spanning language instruction, vision, and objects. We systematically evaluate several state-of-the-art VLA architectures on this suite to understand their generalization capability. Our results show that while VLM backbones endow VLAs with robust perceptual understanding and high‐level planning, which we refer to as good intentions, this does not reliably translate into precise motor execution: when faced with out-of-distribution observations, policies often exhibit coherent intentions, but falter in action execution. Moreover, fine‐tuning on action data can erode the original VLM’s generalist reasoning abilities. We release our task suite and evaluation code to serve as a standardized benchmark for future VLAs and to drive research on closing the perception‐to‐action gap. More information, including the source code, can be found at \url{https://ai4ce.github.io/INT-ACT/}
\end{abstract}

\section{Introduction}
\label{sec:intro}
\input{sections/1_intro}

\section{Related Work}
\label{sec:related}
\input{sections/2_related_work}

\section{\suite{}: A VLA Probing Suite}
\input{sections/3_method}

\section{Experiments and Results}
\label{sec:results}
\input{sections/4_result}

\section{Conclusion and Limitation}
\label{sec:5_conclusion}
\input{sections/5_conclusion}


\bibliography{mybib}
\bibliographystyle{mybib}

\input{sections/appendix}


\end{document}

%% file: sections/1_intro.tex
Building on the significant progress of large Vision-Language Models (VLMs) ~\citep{chen2022pali, liu2023visual, Gemini,OpenAI2024gpt4o} in computer vision and multi-modal tasks, Vision-Language-Action (VLA) ~\citep{openVLA2024, pi02024, pi0_52025} models are emerging to transfer these generalist skills to embodied agents and robotics. 
The ultimate goal is to create robots that can interpret natural language instructions, perceive and plan under complex and diverse environments, and execute versatile behaviors, all under few- or zero-shot scenarios.

This goal calls for a strong generalization capability beyond the training and fine-tuning of robotics data. Early results are encouraging: VLAs can natively follow high‐level language instructions in cluttered scenes and perform highly dexterous manipulation on challenging objects~\cite{openVLA2024, pi02024, pi0_52025}. Moreover, since VLAs are typically pre-trained on a large-scale cross-embodiment dataset \cite{open_x_embodiment_rt_x_2023}, they can be fine-tuned on new tasks with novel configurations for quick adaptation, partially alleviating the data scarcity problem plaguing robotics research.

Despite this progress, there is no consensus on how to systematically measure the generalization capability of VLAs, which is a supposedly strong point of leveraging the modern VLMs over previous perceptual systems. Notably, the current evaluation procedure for VLA generally relies on:
\begin{enumerate}[leftmargin=0.5cm]
    \item Recent simulation benchmarks such as CALVIN \cite{mees2022calvin}, LIBERO \cite{liu2023libero} and
    SimplerEnv \cite{li24simpler} are equipped with natural language instructions, enabling the evaluation of vision-language-action (VLA) models' instruction-following capabilities, which differentiate them from the more traditional benchmarks such as RLBench \cite{james2019rlbench} or RoboMimic \cite{robomimic2021} that focus primarily on low-level control and imitation, with limited or no support for diverse language-conditioned tasks. However, the variety of tasks is relatively limited, and VLA's generalization capability is not targeted.
    \item Real-world robot setup as seen in \cite{openVLA2024, pi02024, pi0_52025}. While this type of evaluation offers endless flexibility and opportunities to test the generalization of a VLA policy, it has considerably bigger barriers than any vision language benchmark due to the time and monetary commitment required.
\end{enumerate}
As a result, it remains unclear to many how well the state-of-the-art VLA models can generalize beyond the robotics datasets they train on by leveraging the remarkable generalization capability of their built-in VLM. To fill this void, we propose a unified probing suite comprising 50 simulation‐based tasks organized into 10 categories that vary systematically along three categories:
\begin{itemize}[noitemsep, topsep=0pt, parsep=0pt, partopsep=0pt, leftmargin=0.5cm]
  \item \textbf{Object diversity:} out-of-distribution (OOD) object, appearance, and affordances unseen during embodiment-finetuning.
  \item \textbf{Language complexity:} from templated commands (e.g., “\texttt{Put A on B}”) to compositional, knowledge and reasoning‐intensive instructions.  
  \item \textbf{Vision-language thinking}: parsing various distractors that are challenging at the perception or planning level.
\end{itemize}

In this work, we make two primary contributions:
\begin{itemize}[leftmargin=0.5cm]
\item We introduce and open-source \suite{}, a comprehensive VLA generalization probing suite comprising 50 tasks across 3 major categories and 10 subcategories, substantially extending the scope of existing VLA benchmarks.
\item Through extensive benchmarking, we uncover two key failure modes in current state-of-the-art VLA models:
\begin{itemize}
\item A persistent and pronounced Intention-Action Gap, where strong semantic understanding under distribution shift fails to translate into reliable execution.
\item Fragile multimodal generalization, particularly under language variations and compounding visual-language distribution shifts.
\end{itemize}
\end{itemize}

%% file: sections/2_related_work.tex
\subsection{Vision-Language-Action Models as Robotic Foundation Models}\label{relate:VLA}

The paradigm of \emph{foundation models}—large, pre-trained architectures fine-tunable for diverse downstream tasks, has transformed natural language processing (NLP) and computer vision (CV). Notable examples include BERT \cite{devlin-etal-2019-bert} and GPT \cite{gpt} for NLP, and CLIP \cite{Radford2021LearningTV} and DINOv2 \cite{oquab2024dinov2} for vision, all showcasing strong zero- and few-shot generalization. This success has shifted research from specialized models towards unified architectures leveraging web-scale pre-training data.

In robotics, \emph{Robotic Foundation Models} aim to equip embodied agents with similar generalist abilities by training on diverse robotics datasets like Open-X Embodiment \cite{open_x_embodiment_rt_x_2023}. Early works, such as RT-1 \cite{rt12022arxiv}, Octo \cite{octo_2023}, and RDT \cite{liu2024rdt}, typically use separate pretrained encoders (e.g., T5 \cite{2020t5} for language, SigLIP \cite{siglip} for vision) to encode multimodal inputs, and subsequently train imitation learning policies (e.g., diffusion policies \cite{chi2024diffusionpolicy}) to map encoded features to robot actions.

Following the success of VLMs~\citep{chen2022pali,chen2023palix,OpenAI2024gpt4o,karamcheti2024prismatic, tong2024cambrian,beyer2024paligemma}, \emph{Vision-Language-Action (VLA)} models aim to integrate the powerful pretrained vision-language models (VLMs) directly, instead of leveraging only pretrained encoders. For instance, RT-2 \cite{rt22023arxiv} fine-tunes PaLM-E \cite{driess2023palme} and PaLI-X \cite{chen2023palix} on combined vision-language and robotics data via autoregressive token prediction, treating discretized robot actions as tokens. OpenVLA \cite{openVLA2024} finetunes Prismatic VLM \cite{karamcheti2024prismatic} on open robotics datasets exclusively, while SpatialVLA \cite{qu2025spatialvla} and FAST \cite{pertsch2025fast} further refine action tokenization strategies. Conversely, models like $\pi_0$ \cite{pi02024} circumvent tokenization by coupling specialized transformers for actions with VLMs (e.g., PaliGemma \cite{beyer2024paligemma}) and applying flow-matching training \cite{liu2022rectifiedflow, lipman2023flowmatching, zhou2024transfusion}.

These models can achieve zero-shot inference through retrieval if the tasks closely resemble training scenarios (e.g., robot, action representation, camera pose, etc). However, fine-tuning on domain-specific demonstration data is necessary for practical deployment. Our study examines VLAs, with the aim of determining whether the pretrained vision-language alignment inherent in VLM contributes to enhanced out-of-distribution generalization after domain-specific fine-tuning.

\subsection{Benchmarks for Vision-Language-Action Models}

Benchmarking in robotics and embodied AI has traditionally relied on standardized simulation environments to systematically evaluate model performance. Early policy learning benchmarks such as RLBench \cite{james2019rlbench}, RoboMimic \cite{robomimic2021}, Factor World \cite{xie2024decomposing}, and RoboManip \cite{RoboManipBaselines_GitHub2024} have predominantly focused on robotic manipulation tasks without explicit language instructions, thus limiting their suitability for evaluating Vision-Language-Action (VLA) models.

Recent benchmarks have begun integrating language instructions. Notable examples include CALVIN \cite{mees2022calvin} and LIBERO \cite{liu2023libero}. However, these benchmarks suffer from simplistic rendering or highly unimodal trajectories, which exacerbates the sim2real gap and undermines their practical applicability.  SimplerEnv \cite{li24simpler} represents a significant advancement by combining system identification, green-screen compositing, and rigorous statistical validation. This methodology effectively ranks policies in a manner closely aligned with their real-world performance.  Nevertheless, SimplerEnv's effectiveness in assessing the broader generalization capabilities essential for VLAs is limited by its restricted set of tasks (fewer than 10 tasks across only two robotic platforms)

Alternatively, real-world benchmarks, recently exemplified by VLA studies \cite{openVLA2024, pi02024, pi0_52025}, offer increased validity through physically diverse and complex test environments. However, these benchmarks entail substantial logistical, financial, and temporal investments, significantly hindering the feasibility and reproducibility of systematic evaluations for the broader research community.

To reconcile these limitations, our work extends SimplerEnv by introducing a comprehensive and unified evaluation suite specifically engineered to probe generalization across three critical dimensions: language complexity, object diversity, and visual variation. This structured approach enables precise assessment of how effectively state-of-the-art VLAs leverage their underlying Vision-Language Model (VLM) capabilities to generalize beyond immediate training and fine-tuning conditions.

%% file: sections/3_method.tex
\label{sec:method}
In this section, we describe our methodology for crafting the \suite{} Suite and the rationale behind our major design choices.


\textbf{Testbed choice}
To minimize barriers in deployment, we choose to build our suite entirely in simulation. Our probing suite extends the SimplerEnv benchmark~\cite{li24simpler}, which is built on the Maniskill2 simulator~\cite{gu2023maniskill2}. It stands out among other VLA benchmarks~\cite{mees2022calvin, liu2023libero} by design to closely mathch models' performance in the real world, making it an ideal testbed for our needs. 
However, the original SimplerEnv benchmark only includes simplistic tasks.
To cater to our need to probe generalization, we significantly expand the original SimplerEnv suite from 4 tasks per dataset to 50 tasks tailored for the BridgeV2 dataset. We also include an additional metric that can keep track of the policy's \textit{intention}. These additions greatly enhance SimplerEnv's breadth and generalization testing potential. 

\textbf{Design Principles} A generalizable VLA policy should inherit VLMs’ strengths in vision-language understanding and broad generalization. We therefore organize our probing tasks into three categories: object diversity, language complexity, and vision-language thinking. We elaborate on each below.

\subsection{Object Diversity}
Truly generalist robot policies require robust perceptual capabilities that extend beyond the specific object distributions encountered during training or fine-tuning. VLMs have demonstrated strong open-vocabulary recognition capabilities \cite{Zohar_2023_CVPR}, enabling reasoning over novel and uncommon visual categories. To assess whether state-of-the-art VLAs can leverage this attribute of VLMs, we significantly expand the object set used in SimplerEnv.

Since it's impossible to audit whether an object appears in the proprietary dataset that $\pi_0$ is pretrained on, we focus on the fine-tune dataset. We argue that this choice also resonates with VLA's real-world deployment: In practice, we care more about whether the VLA fine-tuned on my deployment dataset can generalize beyond the objects shown in my dataset, since additional fine-tuning incurs extra cost.

We begin by auditing all objects present in the BridgeV2 dataset. To introduce out-of-distribution (OOD) diversity, we first include additional household items inspired by other robotic benchmarks such as LIBERO \cite{liu2023libero} and Fractal~\cite{rt12022arxiv}, which feature similar affordances but never appear in the BridgeV2 dataset. Beyond that, we incorporate objects from entirely different domains, such as industrial tools. These additions are deliberately chosen to induce visual mismatch with the pretrained representations learned on domestic data, testing the generalization limits of the VLA policies.

The gathered objects are placed into simple \texttt{PUT \{Source\} ON \{Target\}} tasks by replacing either the source or the target objects with an OOD object.
Then, we expand the combinatorial complexity of the tasks by making both the source object and the target object OOD. 
Additionally, to probe whether the VLA policy falls to memorizing spurious correlations from training data, we introduce OOD Relations between objects seen in the BridgeV2 dataset, such as putting a carrot on a sponge.

\subsection{Language Complexity}
Recent advancements in Vision-Language Models (VLMs)~\cite{openai2024gpt4ocard, tong2024metamorph} demonstrate impressive capabilities in understanding diverse and semantically rich instructions such as those that require commonsense and reasoning. For instance, when prompted with ``\textit{Create an image of the flower celebrated in spring festivals in the country where sushi originated,}'' modern VLMs correctly infer Japan as the country and subsequently cherry blossom as the flower.

In contrast, recent robotics datasets and benchmarks generally rely on short, templated commands without ambiguity, such as \texttt{PUT \{X\} ON \{Y\}}. To probe whether VLAs inherit the adavanced generalization abilities of their underlying VLMs, we sweep through all instructions in the BridgeV2 dataset and augment the original SimplerEnv instructions with more complex linguistic variations, including changes in action verbs, semantic negation, referential appearance, and commonsense cues. Examples of such variations are provided in Table~\ref{fig:language_variation}. The variation logic are as follows:
\begin{itemize}[noitemsep, topsep=0pt, parsep=0pt, partopsep=0pt, leftmargin=0.5cm]
    \item Language Action: paraphrasing verbs to be compositional and less frequent in BridgeV2.
    \item Language Negation: adding negation such as \texttt{not}, \texttt{don't} to irrelevant objects.
    \item Language Appearance: replacing object with descriptive words, such as \texttt{a purple object}, instead of \texttt{an eggplant}.
\end{itemize}

\begin{figure}[t]
    \centering
    \includegraphics[width=\linewidth]{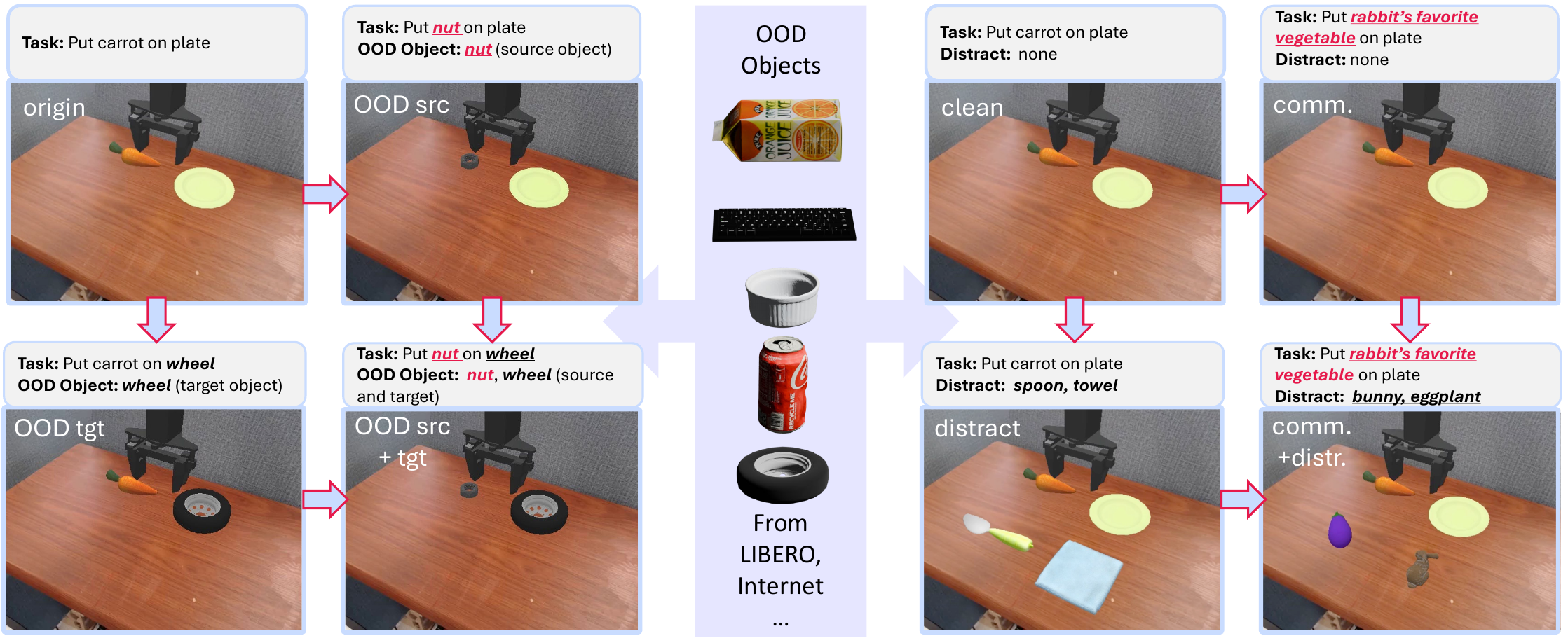}
    \caption{\textbf{Left}: Examples of tasks with out-of-distribution objects. \textbf{Right}: Examples of tasks with commonsense reasoning, distractors, and commonsense reasoning + distractors.}
    \label{fig:object_vision}
    \vspace{-10pt}
\end{figure}

\subsection{Vision-Language Thinking}
Robust real-world deployment of VLA models requires resilience to the complex and cluttered visual environments that typify human environments. However, the original SimplerEnv benchmark focuses on minimalist scenes that include only the source and target objects relevant to a given task. 
To address this limitation, we augment the benchmark with visually richer scenes by introducing additional objects that are irrelevant to the specified task. These objects are chosen from a broader pool of distractors and placed throughout the scene to simulate clutter and occlusion. Beyond generic clutter, we also include semantically challenging distractors that require commonsense reasoning to disambiguate. For example, in a task where the agent is instructed to pick up an orange juice box, we introduce an orange nearby. This setup probes whether the VLA model can resolve ambiguity based on language context and semantic cues.

Prior research \cite{li24simpler, xie2024decomposing} has shown that visual factors such as lighting changes, background variation, and camera pose shifts can degrade model performance. Unfortunately, SimplerEnv's reliance on green-screen compositing complicates the simulation of lighting, texture, and camera pose variation. However, we note that recent studies have found that state-of-the-art imitation learning and VLA models are relatively robust to lighting and background perturbations, especially when pretrained with diverse visual datasets \cite{xie2024decomposing}. That said, variations in camera pose remain a critical challenge. We refer our readers to previous research \cite{li24simpler, xie2024decomposing} on the impact of these visual factors on large-scale robot learning policy.

%% file: sections/4_result.tex

\subsection{Model Selection and Training Protocol}

We evaluate four prominent models and their variants in \suite{} suite: $\pi_0$ \cite{pi02024}, SpatialVLA \cite{qu2025spatialvla}, Magma \cite{yang2025magma}, and Octo \cite{octo_2023}. They are selected to represent a diverse set of architectural paradigms and training methodologies in the current VLA landscape. All finetuning experiments are conducted on BridgeV2, a popular large robot learning dataset consistent with SimplerEnv. We adhere to the training and fine-tuning protocols in the respective papers to ensure fair and consistent comparisons.

\textbf{Pi-Zero} ($\boldsymbol{\pi}_{\mathbf{0}}$), is chosen for its consistently strong performance across standard VLA benchmarks and serves as a representative of the diffusion/flow-based modeling family. $\pi_0$ is evaluated in two variants: (1) fine-tuned from the released checkpoint on BridgeV2 (denoted \textit{$\pi_0$-finetune}), which mirrors the intended usage, and (2) trained from scratch using their pretrained VLM (i.e., Paligemma) and only BridgeV2 as the robot data source (denoted \textit{$\pi_0$-scratch}), which serves as a strong baseline for multi-task imitation learning.

\textbf{SpatialVLA} is included as a leading autoregressive VLA model, frequently reported to outperform other autoregressive counterparts such as OpenVLA in various experimental settings \cite{qu2025spatialvla, simplerEnvIssue78}. SpatialVLA is used directly from an official BridgeV2-finetuned checkpoint.

\textbf{Magma}, while built on similar autoregressive models, leverages a co-training regime with vision-language data, a strategy that retains model the vision-language understanding capability for VQA and agentic tasks. We are curious to know whether this co-training can also help its generalization in action, particularly under our proposed out-of-distribution evaluations. Magma is evaluated in a zero-shot configuration, following the protocol in its paper.

\textbf{Octo} is trained on cross-embodiment datasets conditioned on both visual and language inputs. Though not a VLA model in the strictest definition discussed in Sec.~\ref{relate:VLA}, it has served as an exemplar large multi-task imitation learning baseline in the VLA literature. We evaluate both Octo Small (27M parameters) and Octo Base (93M parameters) variants. Octo is run in zero-shot mode. Although fine-tuning is recommended for novel scenes, our preliminary trials, as well as independent research \cite{qu2025spatialvla}, suggest that zero-shot Octo actually performs slightly better in BridgeV2 scenes. Therefore, we stick to the zero-shot setting adhere to the literature.



\begin{figure}[t]
    \centering
    \scalebox{0.8}{
    \begin{subfigure}[b]{0.45\textwidth}
        \centering
        \includegraphics[width=0.9\textwidth]{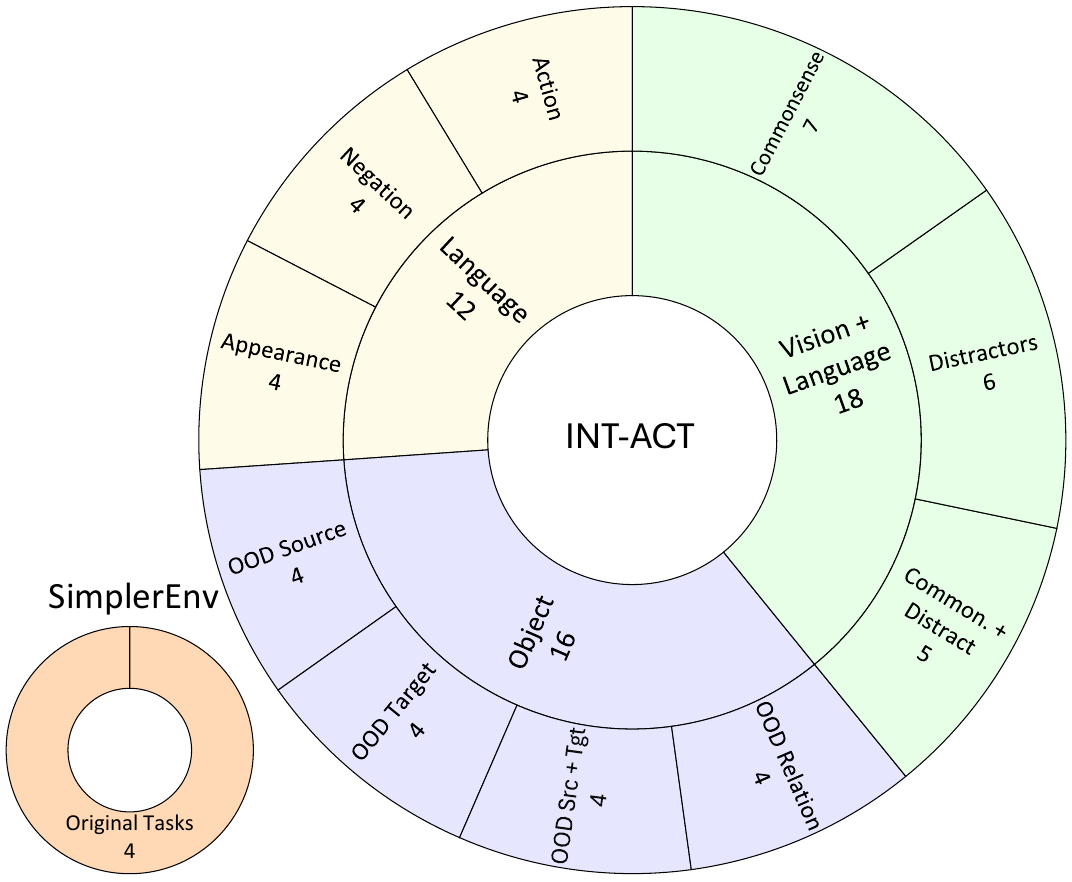}
        \caption{The distribution of our tasks compared to SimplerEnv's original tasks}
        \label{fig:task_distribution}
    \end{subfigure}
    \begin{subfigure}[b]{0.45\textwidth}
        \centering
        \includegraphics[width=0.9\textwidth]{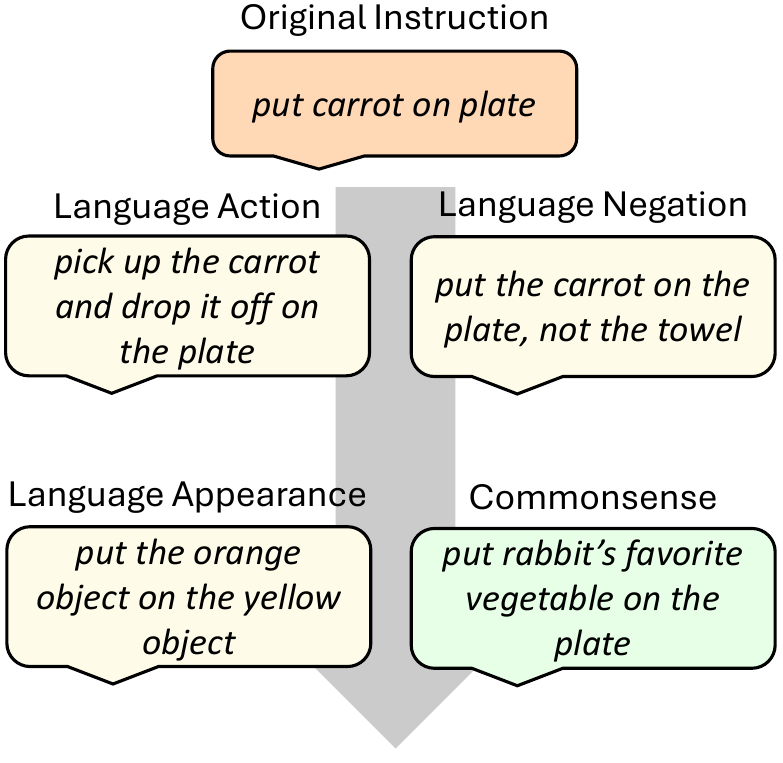}
        \caption{Sample of language variations}
        \label{fig:language_variation}
    \end{subfigure}
    }
    \caption{
    Illustration of the \suite{} probing suite. \textbf{Left}: suite category breakdown. \textbf{Right}: illustraion of language variations.
    }
    \label{fig:ood_results}
    \vspace{-0.5cm}
\end{figure}

Our benchmark is organized into three major evaluation categories of object, language and vision-language, each with more subcategories as described in Sec. \ref{sec:method}. Together with the original SimplerEnv tasks, our suite consists of 50 tasks designed to provide comprehensive and fine-grained evaluation. A breakdown of the task distribution is provided in Fig~\ref{fig:task_distribution}.


\textbf{Evaluation Procedure} For each task, we evaluate each model across 24 episodes, corresponding to all possible scene and object configurations predefined by ManiSkill2. Each configuration is repeated across 3 random seeds, following standard practice in prior works \cite{octo_2023, openVLA2024}. All reported metrics are averaged across episodes and seeds.

\textbf{Metrics} We report three primary metrics:
\begin{itemize}[noitemsep, topsep=0pt, parsep=0pt, partopsep=0pt, leftmargin=0.5cm]
\item \textbf{Grasp Success Rate}: Whether the robot's gripper successfully grasps the correct source object at any point during the episode. This metric is natively supported by SimplerEnv.
\item \textbf{Intention Correct Rate}: A new metric introduced in our benchmark, defined by whether the gripper moves to within a small radius of the correct source object at any frame. This captures the policy's intention to grasp the correct object, even if the grasping fails afterwards.
\item \textbf{Task Success Rate}: If the task is successfully completed, natively supported by SimplerEnv.
\end{itemize}

These metrics jointly capture both motor-level competence (grasping) and semantic grounding (correct object targeting), enabling a finer evaluation of policy behavior, decomposing the perception and action stages.

\subsection{Performance analysis}
\begin{figure}[ht]
    \centering
    \includegraphics[width=\linewidth]{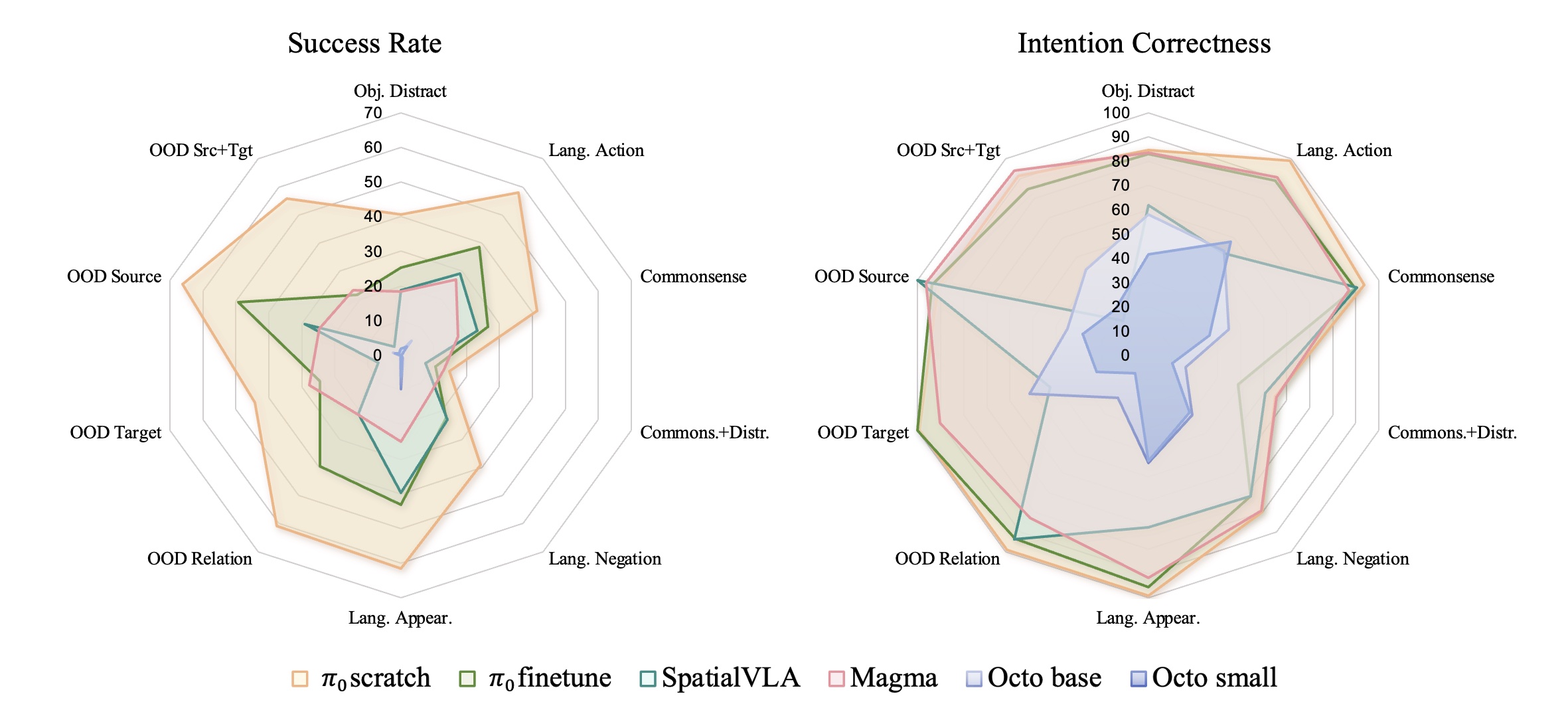}
    \caption{\textbf{The Intention-Action Gap} illustrated by comparing the two radar maps. Task Success Radar on the left, Intention Correctness Radar on the left. Best viewed in color.}
    \label{fig:radar}
\end{figure}

\textbf{Intention-Action Gap}
Table~\ref{tab:main} summarizes average performance across generalization categories, with radar plots in Figure~\ref{fig:radar} visualizing the Intention Correctness and Task Success Rate. A pronounced gap emerges: while most VLM-initialized VLAs achieve near-perfect Intention Correctness ($80-100\%$) across all categories—even weaker Octo variants show $50\%+$, their Task Success Rates drop drastically. The $\pi_0$ scratch model leads performance but still falls far short of matching its intention correctness, highlighting a stark intention-to-execution disparity.
This observation confirms that VLMs endow VLA policies with a generalizable notion of “what to do,” yet executing that intent—especially under distributional shifts, remains challenging. This Intention-Action Gap persists even when comparing VLM-based VLAs against non-VLM baselines like Octo, suggesting the limitation lies not in semantic grounding, but in translating it into robust low-level control.
To deepen understanding, we analyze model behaviors using \suite{} with three major questions.
The full results are attached in the Appendix~\ref{app:full_results}.

\begin{table}[t]
    \centering
    \vspace{1em}
    \caption{Results of the \suite{} probing suite by category.}
    \fontsize{5pt}{4.8pt}\selectfont
    \setlength\tabcolsep{3pt}
    \scalebox{1.57}{
    \begin{tabular}{cc|c|ccccccccccc}
        & & &
        \rotatebox{90}{Original} &
        \rotatebox{90}{OOD Source} &
        \rotatebox{90}{OOD Target} &
        \rotatebox{90}{OOD Src + Tgt} &
        \rotatebox{90}{OOD Relation} &
        \rotatebox{90}{Lang. Action} &
        \rotatebox{90}{Lang. Negation} &
        \rotatebox{90}{Lang. Appear.} &
        \rotatebox{90}{Obj. Distract} &
        \rotatebox{90}{Commons.} &
        \rotatebox{90}{Commons.+Distr.} \\
          \midrule
        Methods & Metric & Avg. & 
        \multicolumn{1}{c}{\cellcolor{orange!10}Simpler} & 
        \multicolumn{4}{c}{\cellcolor{blue!10}Object OOD} & 
        \multicolumn{3}{c}{\cellcolor{yellow!10}Language} & 
        \multicolumn{3}{c}{\cellcolor{green!10}Vision + Lang.} \\
        \midrule
        \multirow{3}{*}{$\pi_{\scriptscriptstyle 0}$ fintune~\cite{pi02024}} 
        & intention & 84.5 & 99.0 & 93.8 & 100  & 84.7 & 93.4 & 88.9 & 71.9 & 95.5 & 82.9 & 89.5 & 38.9 \\
        & grasp     & 54.4 & 77.8 & 64.2 & 66.1 & 56.9 & 61.1 & 63.9 & 44.4 & 56.3 & 61.1 & 45.4 & 17.2 \\
        & success   & 30.4 & 47.6 & 49.3 & 24.4 & 21.5 & 39.6 & 38.5 & 22.2 & 43.1 & 25.2 & 26.4 & 10.6 \\
        \midrule
        \multirow{3}{*}{$\pi_{\scriptscriptstyle 0}$ scratch~\cite{pi02024}} 
        & intention & 89.5 & 100  & 91.3 & 99.2 & 91.3 & 99.0 & 99.3 & 79.9 & 99.0 & 84.7 & 93.7 & 55.0 \\
        & grasp     & 66.7 & 87.2 & 77.4 & 71.1 & 88.2 & 81.6 & 84.7 & 59.4 & 69.1 & 61.8 & 59.5 & 26.9 \\
        & success   & 48.9 & 68.1 & 66.3 & 44.2 & 75.7 & 60.8 & 58.0 & 39.2 & 61.5 & 25.2 & 41.3 & 14.7 \\
        \midrule
        \multirow{3}{*}{Magma~\cite{yang2025magma}} 
        & intention & 85.4 & 96.5 & 96.5 & 90.3 & 94.1 & 83.0 & 90.6 & 79.2 & 91.7 & 83.6 & 87.1 & 55.6 \\
        & grasp     & 46.5 & 57.6 & 58.3 & 46.9 & 54.5 & 50.3 & 56.6 & 37.5 & 45.8 & 41.2 & 45.0 & 25.8 \\
        & success   & 21.6 & 29.2 & 24.7 & 27.8 & 23.3 & 21.2 & 27.1 & 14.2 & 25.0 & 18.3 & 17.5 & 13.1 \\
        \midrule
        \multirow{3}{*}{SpatialVLA~\cite{qu2025spatialvla}} 
        & intention & 69.6 & 100  & 100  & 42.5 & 17.7 & 93.8 & 52.1 & 71.9 & 70.8 & 61.8 & 90.5 & 50.8 \\
        & grasp     & 41.3 & 59.4 & 54.2 & 26.7 & 12.5 & 43.8 & 38.5 & 40.6 & 58.3 & 43.1 & 48.2 & 25.0 \\
        & success   & 21.5 & 39.6 & 29.2 & 6.7  & 3.1  & 20.8 & 29.2 & 22.9 & 39.6 & 18.7 & 23.2 & 7.5 \\
        \midrule
        \multirow{3}{*}{Octo Small~\cite{octo_2023}} 
        & intention & 31.5 & 56.6 & 28.5 & 22.5 & 23.3 & 9.0 & 58.0 & 29.2 & 43.4 & 41.7 & 26.6 & 10.3 \\
        & grasp     & 13.9 & 26.7 & 14.6 & 10.3 & 13.2 & 3.1 & 30.6 & 12.2 & 16.0 & 19.4 & 9.1  & 1.9 \\
        & success   & 1.6  & 5.2  &  1.7 & 0.0  & 0.7  & 0.0 & 3.1  & 0.7  & 5.9  & 1.9  & 0.2  & 0.0 \\
        \midrule
        \multirow{3}{*}{Octo Base~\cite{octo_2023}} 
        & intention & 41.5 & 74.7 & 35.1 & 51.4 & 43.7 & 21.5 & 53.1 & 30.6 & 44.4 & 58.1 & 34.9 & 16.1 \\
        & grasp     & 14.1 & 27.8 & 13.2 & 15.0 & 17.0 & 6.2  & 21.2 & 7.3  & 17.7 & 19.0 & 10.7 & 4.4 \\
        & success   & 2.4  & 8.3  & 2.4  & 0.3  & 0.3  & 0.7  & 5.2  & 1.0  & 9.7  & 1.2  & 0.0  & 0.0 \\
        \bottomrule
    \end{tabular}
    }
    \label{tab:main}
    \vspace{-10pt}
\end{table}

\begin{figure}
    \centering
    \begin{subfigure}[b]{0.49\textwidth}
        \centering
        \includegraphics[width=\textwidth]{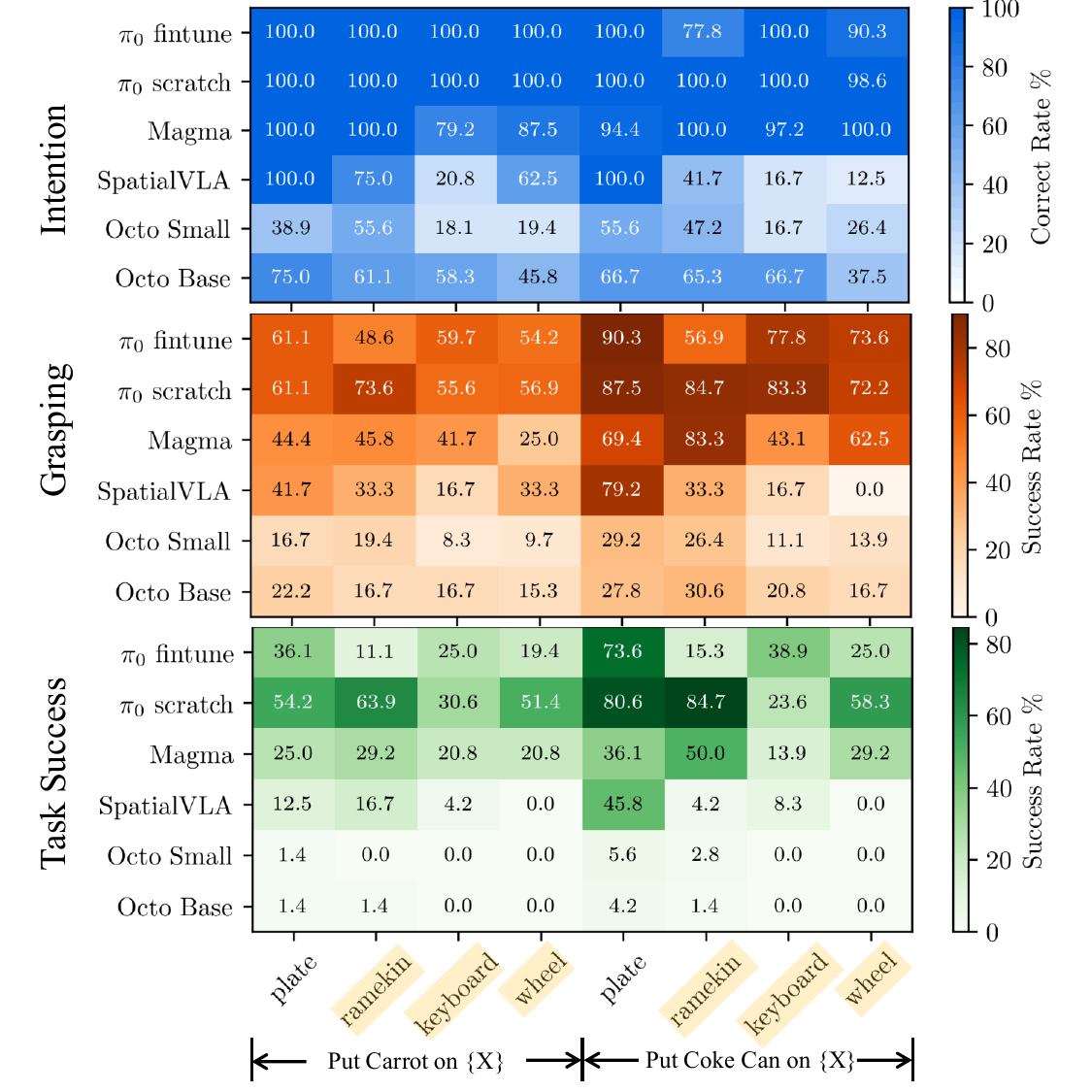}
        \caption{OOD target objects.}
        \label{fig:heatmap_ood}
    \end{subfigure}
    \begin{subfigure}[b]{0.49\textwidth}
        \centering
        \includegraphics[width=\textwidth]{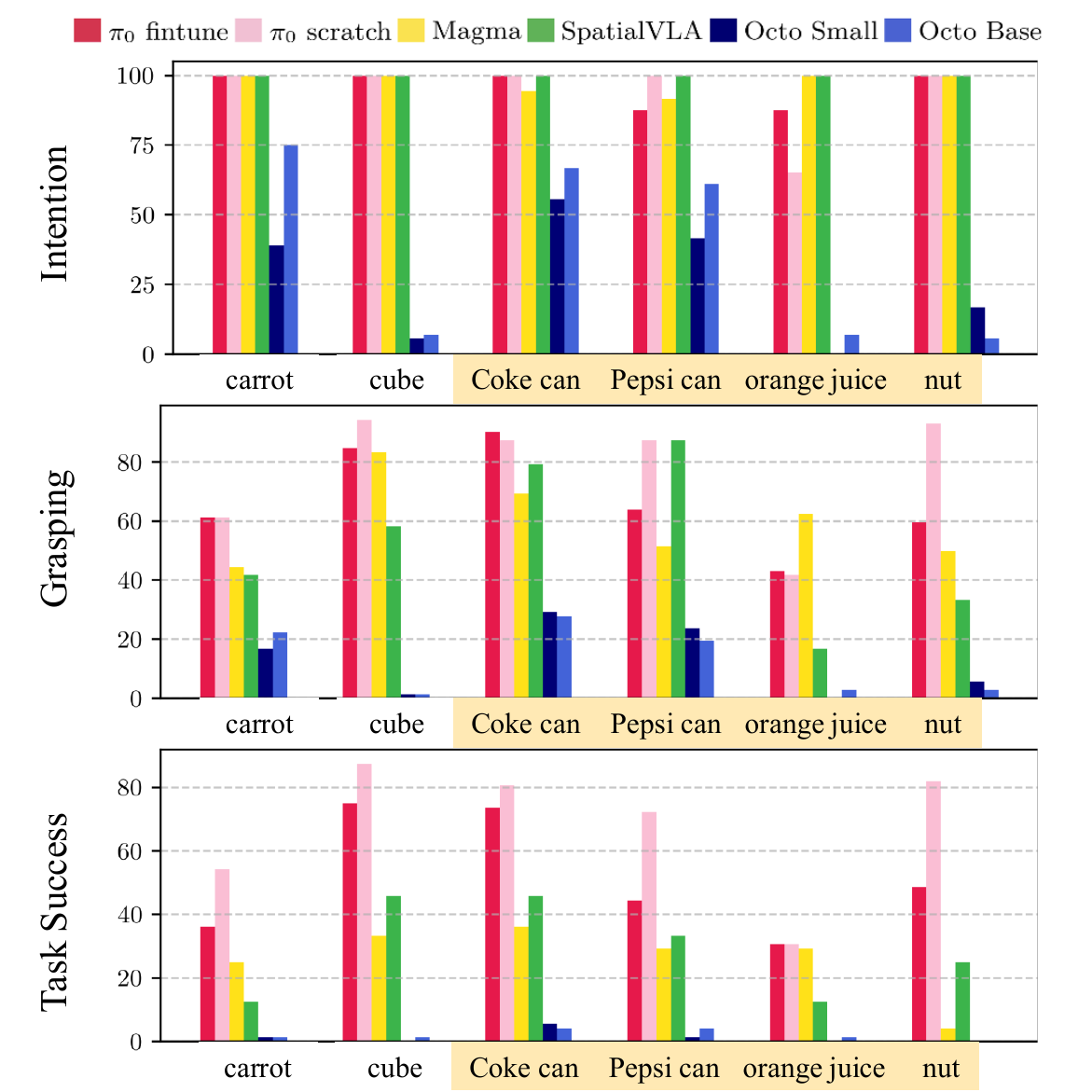}
        \caption{OOD source objects.}
        \label{fig:bar_ood}
    \end{subfigure}
    \caption{OOD generalization results. Out-of-Distribution objects are painted in \colorbox{lightorange}{orange}. 
    }
    \label{fig:ood_results}
\end{figure}

\question{1}{How do VLAs generalize to OOD objects?}
\textbf{VLAs show robust intention to OOD objects, but execution falls short. Changing the target can affect grasping of the source.}
We break down per-object OOD performance in Figure \ref{fig:ood_results}. Specifically, Figure~\ref{fig:heatmap_ood} presents metrics for putting two source objects, carrot and coke can (OOD), on four target objects, plate, ramekin, keyboard and wheel, respectively. All targets except plate are OOD targets. 
Consistent with the overall performance, all VLA models demonstrate strong intention correctness across every target. By contrast, grasping success rate varies significantly.
Interestingly, even when the source object is kept unchanged and only the target object varies, the grasping success rate can still swing up to 40\% in some cases. Since the underlying grasp primitive should be identical, this suggests the brittle coupling of high-level perception and planning with low-level action of these end-to-end policies.

For comparisons on OOD source objects in Figure~\ref{fig:bar_ood}, we see that SpatialVLA, Magma, and $\pi_0$ models all demonstrate robust intentions. Meanwhile, \textbf{grasping and task success rates vary more by object identity than by their OOD status}. This is also reflected in the heatmaps in Figure~\ref{fig:heatmap_ood}. The coke can, though unseen in BridgeV2, generally yields higher grasping success rates than the carrot, a common object in BridgeV2.
hese per-object OOD analyses highlight an \textit{Intention-Action gap}: the generalized perception capabilities of VLMs do not directly translate to generalized action performance. This observation suggests a need for improved architectural designs for integrating VLMs into robotic systems, possibly by incorporating insights from grounding or modularized approaches such as~\cite{huang2024rekep,team2025gemini-robotics, wang2024seedo}.

\question{2}{How well do VLAs preserve the linguistic capability?}
\textbf{VLAs do not preserve the linguistic capability of the underlying VLM when facing complex language instruction.} We list performance delta in three categories of language variations in Table~\ref{tab:lang_variation}. Both \textit{Lang. Action} and \textit{Lang. Negation} emphasizes the linguistic robustness while \textit{Lang. Appearance} additionally involves simple visual understanding.

Results show all models suffer performance drops in all categories. 
To verify if simply paraphrasing task instructions with more diverse vocabulary will mitigate the challenge, we additionally fine-tuned a $\pi_0$ checkpoint with task paraphrasing, denoted as $\pi_0$ finetune+\textit{rephrase}. Results show that although paraphrasing improves over $\pi_0$ finetune on various tasks in general, it still degrades severely in the face of language variation. One exception is the intention in \textit{Lang. Action}, which gets restored by finetuning with paraphrasing, but this does not translate into more robust action execution.
We conclude that linguistic capabilities from VLMs are not fully preserved after end-to-end VLA training; execution still breaks under simple language perturbations.
This comes as a surprise since the backbone VLMs, such as Paligemma and Llama3, are popular choices in vision-language tasks, so one would expect their generalizability in language can be easily absorbed by the VLAs. Magma, possibly thanks to its joint vision-language co-training, shows best robustness to action words variations as seen in the \textit{Lang. Action} category. Nonetheless, the loss of linguistic generalizability after VLA training revealed by \suite{} is significant.

\definecolor{darkgreen}{rgb}{0.0, 0.5, 0.0}
\definecolor{oai-gray-300}{HTML}{E5E5E5}
\newcommand{\plusvalue}[1]{\hspace{0.3em}\textcolor{darkgreen}{{\fontsize{7pt}{8pt}\selectfont\textbf{(+#1)}}}}
\newcommand{\minusvalue}[1]{\hspace{0.3em}\textcolor{red}{{\fontsize{7pt}{8pt}\selectfont\textbf{(-#1)}}}}
\begin{table}[t!]
    \centering
    \caption{
\textbf{Impact of language variation on VLA generalization.}
Average performance on tasks before applying the language variations is listed in \colorbox{oai-gray-300}{light gray}.
}
    \fontsize{7.5pt}{8.0pt}\selectfont
    \setlength\tabcolsep{3.0pt} 
    \scalebox{0.92}{
    \begin{tabular}{cr|lll|lll|lll  }
    \rowcolor{oai-gray-300}($\Delta$ perf.) & &
     \multicolumn{3}{c|}{\textbf{+ Lang. Action}} &
     \multicolumn{3}{c|}{\textbf{+ Lang. Appearance}} &
     \multicolumn{3}{c}{\textbf{+ Lang. Negation}} \\
      Method & &
\rotatebox{0}{Intention} &
\rotatebox{0}{Grasp} &
\rotatebox{0}{Success} &
\rotatebox{0}{Intention} &
\rotatebox{0}{Grasp} &
\rotatebox{0}{Success} &
\rotatebox{0}{Intention} &
\rotatebox{0}{Grasp} &
\rotatebox{0}{Success} \\
      \midrule 
\multirow{2}{*}{SpatialVLA} &  \cellcolor{oai-gray-300}{before} & \color{gray!70}100 &\color{gray!70}59.4 &\color{gray!70}39.6 & \color{gray!70}100 & \color{gray!70}59.3& \color{gray!70} 43.8 &\color{gray!70} 100 & \color{gray!70}43.7 & \color{gray!70} 22.9\\  
 & after & 52.1\minusvalue{47.9} & 38.5\minusvalue{20.8} & 29.2\minusvalue{10.4} & 70.8\minusvalue{29.2} & 58.3\minusvalue{1.0} & 39.6\minusvalue{4.2} & 71.9\minusvalue{28.1} & 40.6\minusvalue{3.1} & 22.9\plusvalue{0.0} \\
 \multirow{2}{*}{Magma} &  \cellcolor{oai-gray-300}{before} & \color{gray!70} 96.5 & \color{gray!70}57.6 & \color{gray!70}29.2 & \color{gray!70}98.0 & \color{gray!70}64.2 & \color{gray!70} 35.1& \color{gray!70} 96.6& \color{gray!70}55.6 & \color{gray!70} 27.7\\
 & after & 90.6\minusvalue{5.9} & 56.6\minusvalue{1.0} & 27.1\minusvalue{2.1} & 91.7\minusvalue{6.3} & 45.8\minusvalue{18.4} & 25.0\minusvalue{10.1} & 79.2\minusvalue{17.4} & 37.5\minusvalue{18.1} & 14.2\minusvalue{13.5} \\
 \multirow{2}{*}{$\pi_0$ scratch} &  \cellcolor{oai-gray-300}{before} & \color{gray!70}100 &\color{gray!70}87.2 & \color{gray!70}68.1 & \color{gray!70}100& \color{gray!70}87.5& \color{gray!70}79.9 & \color{gray!70}91.4 & \color{gray!70}71.6 & \color{gray!70}63.2\\
  & after & 99.3\minusvalue{0.7} & 84.7\minusvalue{2.4} & 58.0\minusvalue{10.1} & 99.0\minusvalue{1.0} & 69.1\minusvalue{18.4} & 61.5\minusvalue{18.4} & 79.9\minusvalue{11.5} & 59.4\minusvalue{12.2} & 39.2\minusvalue{24.0} \\
 \multirow{2}{*}{$\pi_0$ finetune} &  \cellcolor{oai-gray-300}{before} & \color{gray!70}99.0 &\color{gray!70}77.8 & \color{gray!70}47.6 & \color{gray!70}99.0& \color{gray!70}77.5& \color{gray!70}59.8 & \color{gray!70}95.9 & \color{gray!70}65.2 & \color{gray!70}46.5\\
  & after & 88.9\minusvalue{10.1} & 63.9\minusvalue{13.9} & 38.5\minusvalue{9.0} & 95.5\minusvalue{3.5} & 56.3\minusvalue{21.2} & 43.1\minusvalue{16.7} & 71.9\minusvalue{24.0} & 44.4\minusvalue{20.8} & 22.2\minusvalue{24.3} \\
  \multirow{2}{*}{\shortstack{$\pi_0$ finetune \\ \textit{+rephrase}}} &  \cellcolor{oai-gray-300}{before} & \color{gray!70}98.6 & \color{gray!70}77.1 & \color{gray!70} 56.6 & \color{gray!70}98.6 & \color{gray!70}81.9& \color{gray!70}70.5 & \color{gray!70}96.9 & \color{gray!70}71.9& \color{gray!70}55.2 \\  
   & after & 98.6\plusvalue{0.0} & 74.3\minusvalue{2.8} & 45.1\minusvalue{11.5} & 90.6\minusvalue{8.0} & 51.0\minusvalue{30.9} & 41.3\minusvalue{29.2} & 78.1\minusvalue{18.8} & 53.1\minusvalue{18.8} & 33.0\minusvalue{22.2} \\
   \bottomrule
    \end{tabular}
}
\label{tab:lang_variation}
\end{table}

\begin{figure}
    \centering
    \begin{subfigure}[b]{\textwidth}
        \centering
        \includegraphics[width=\textwidth]{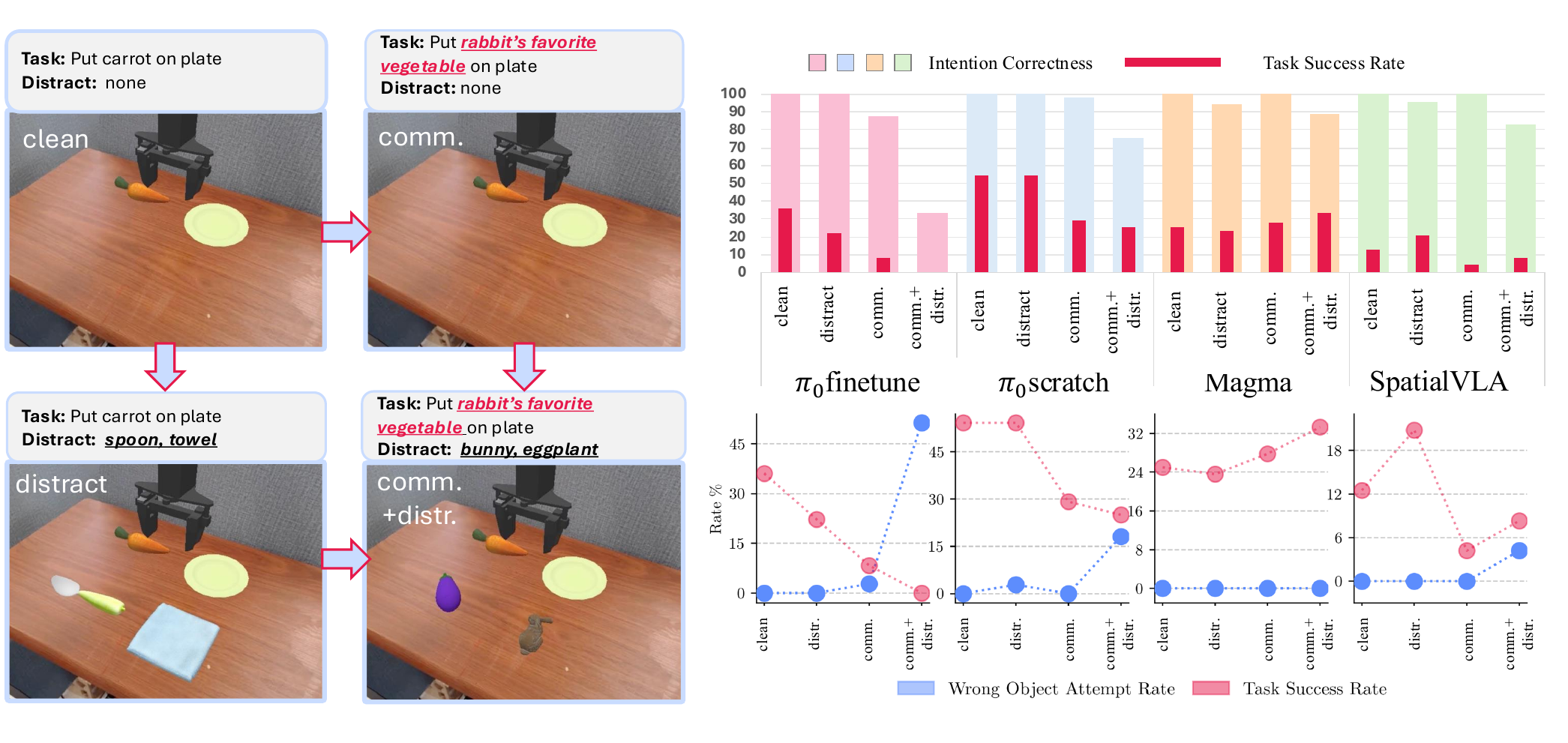}
        \caption{Carrot case study.}
        \label{fig:distract_carrot}
    \end{subfigure}
    \vspace{0.5em}
    \begin{subfigure}[b]{\textwidth}
        \centering
        \includegraphics[width=\textwidth]{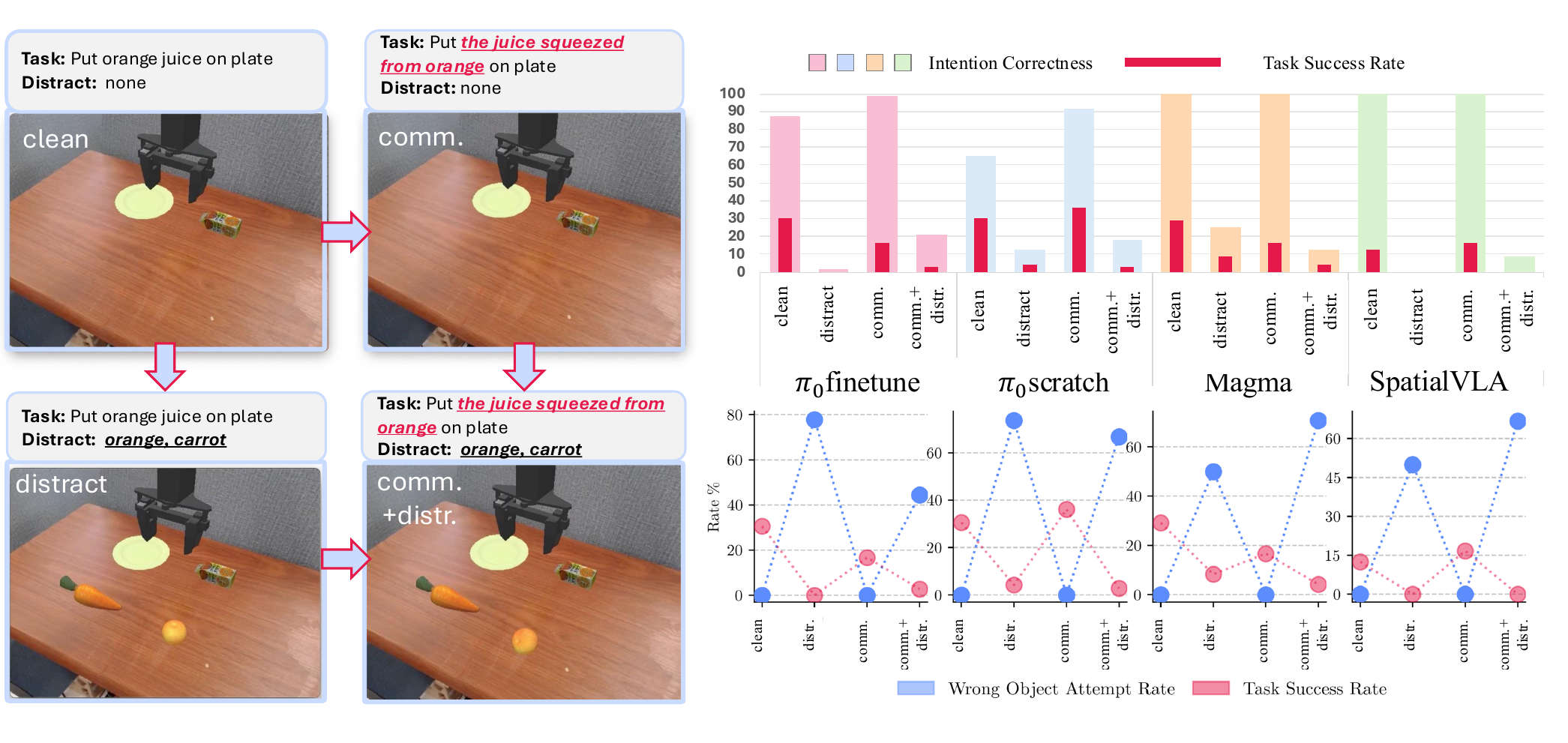}
        \caption{Orange juice case study.}
        \label{fig:distract_juice}
    \end{subfigure}
    \caption{Case studies showing the impact of visual distractions and language commonsense variations. Task illustrations are on the left. Success rate, Intention Correctness and Wrong Object Attemp Rate are grouped by models on the right.}
    \label{fig:distract_case_studies}
\end{figure}

\question{3}{How well do VLAs preserve the vision-language thinking?}
Lastly, we further test the robustness of VLAs in scenarios involving commonsense reasoning and visual distractions. To this end, we analyze the performance trend in three progressively complex categories: \textit{Distraction}, \textit{Language Commonsense}, and \textit{Language Commonsense with Distraction}.
We take two specific tasks for case studies: \texttt{Put carrot on plate} and \texttt{Put orange juice on plate}, and extend them by adding distraction objects and replacing the source object with commonsense paraphrasing as illustrated in Figure~\ref{fig:distract_case_studies}.
In addition to Intention Correctness and Task Success Rate, we additionally monitor if any non-source object is moved during the episode and report it as the \textit{Wrong Object Attempt Rate}.

For the carrot case, all models maintain high intention correctness with either distractions or commonsense paraphrases alone, despite the decreased success rate. Wrong object attempts stay near zero in these single‐factor settings. However, when the toy bunny is added to the scene, together with the language commonsense factor "rabbit's favorite vegetables", the wrong object attempt surges, especially for $\pi_0$ finetune, indicating a breakdown in correctly binding linguistic intent to the visual scene.
For the orange juice case, we take advantage of the name and add an orange as the distraction, which causes a language distraction too. This completely breaks the models, yielding high wrong object attempt rates across all of them.
In conclusion, the probing suggests that \textbf{while VLAs have some robustness to isolated commonsense and visual distractions, they are brittle under multimodal ambiguity}, where linguistic priors override visual grounding, leading to systematic misbehavior.

To further support our observations in Questions 2 and 3, we prompted the pretrained PaliGemma by with instructions and objects for a simple VQA, as shown in Figures~\ref{fig:paligemma_prompt} and~\ref{fig:paligemma_vqa} in Appendix~\ref{app:prompt}. The results indicate that even as a smaller language model, and not specifically optimized for instruction following, PaliGemma demonstrates considerable understanding before being fine-tuned as a Visual Language Action (VLA) model. This once again suggests that there's significant room to improve VLA architecture design to better leverage the inherent capabilities of VLMs, rather than disrupting them. Insights from paradigms like freezing VLMs in image generation, as explored in~\cite{pan2025metaquery}, could prove beneficial here.

%% file: sections/5_conclusion.tex
In this work, we proposed a VLA generalization probing suite \suite{} with 50 tasks across 10 categories in object, language, and vision, aiming to understand the generalization boundary of state-of-the-art VLAs. Our benchmarking effort using \suite{} reveals two critical challenges in the current generation of VLA models.

First is a persistent and wide Intention-Action Gap. While VLM backbones confer impressive semantic understanding, enabling VLAs to generalize high-level goals across distribution shifts, this competence rarely transfers to reliable execution. Grasp success and task completion suffer dramatically, even in tasks with correct intent.

Further probing shows that VLA models degrade under simple language variations, failing to preserve the generalization capabilities of the underlying language backbones. Multimodal reasoning during action execution, especially when vision and language out-of-distribution compounds, remains a significant weakness. Even models with joint vision-language pretraining and language augmentation fine-tuning struggle with commonsense disambiguation in the presence of distractions.

Our work is not without limitations. The current \suite{} is based on the BridgeV2 dataset and its associated robot and sensor configurations. Expanding it to more robot embodiments would make the results more compelling. At the same time, since the possible objects and language distribution shift is theoretically infinite, it would be great if we could incorporate technology like LLMs or 3D generative models to automate the task creation process to increase the number of significantly out-of-distribution tasks, thus providing more compelling statistical signals.
In addition, \suite{} is developed from SimplerEnv so it is fully in simulation. This makes it extremely accessible but also inherits the same sim2real gap. Although SimplerEnv is built to match real-world performance, extending the \suite{} to real-world is still a worthwhile future step. 

\paragraph{Acknowledgement.}
Irving Fang and Juexiao Zhang share the first authorship. Their collaboration initiated the project. They both conducted experiments and wrote the paper.
Juexiao Zhang led the project, designed and implemented the evaluation metric and tasks.
Irving Fang suggested emphasize the project on evaluation and implemented the code infrastructure.
Shengbang Tong engaged in conceptualization, advised on benchmarking, VLM and VLA training, and paper writing.
Chen Feng oversaw and advised the project.
We thank Saining Xie and Yiming Li for their helpful discussions, SimplerEnv authors and Allen Z Ren for opensourcing their code as our important reference.
The work was supported in part through NSF grants 2238968, 2322242 and 2026479, and the NYU IT High Performance Computing resources, services, and staff expertise.

%% file: sections/appendix.tex
\newpage
\appendix

\section{Experiment details}
\label{app:details}

\textbf{Training $\pi_0$ on BridgeV2}.
When fine-tuning and training $\pi_0$ on the BridgeV2 dataset, we employ 4 H100 80GB GPUs from Nvidia and the following parameters. The parameters are entirely the same for both training and fine-tuning. It takes about 17 hours to finish the training loop for 15 epochs on BridgeV2.
\begin{table}[h]
    \centering
    \caption{Hyperparameters for training and fine-tuning $\pi_0$}
    \begin{tabular}{cc}
    \toprule
    Hyperparameters  & Value \\
    \midrule
    Global Batch Size     &  1024 \\
    Epochs & 15 \\
    Single Card Batch Size & 32 \\
    Optimizer LR & $5e-5$\\
    Optimizer Betas & [
        0.9,
        0.999
    ]\\
    Optimizer Eps & 1e-8\\
    Optimizer Weight Decay & 0\\
    Scheduler Warmup Steps & 200\\
    Chunk Size & 4\\
    Freeze Vision Encoder & False\\
    Language Tokenizer Max Length & 72\\
    MLP Projector Width & 1024\\
    \bottomrule
    \end{tabular}
    \label{tab:my_label}
\end{table}

Notably, the max state and action dimensions for the MLP projectors are set differently for fine-tuning and training from scratch. Specifically, the fine-tuning version uses 32 because the official $\pi_0$ checkpoint uses 32 to accommodate various embodiments in the dataset. However, for training from scratch, we discover that setting the max state and action dimension to exactly the dataset's state and action dimension (7 in this case), thus avoiding 0-padding, does improve performance. This configuration follows another third-party open-source implementation of $\pi_0$\footnote{https://github.com/allenzren/open-pi-zero}.

\textbf{Evaluation $\pi_0$ variants}.
When evaluating $\pi_0$ finetune and $\pi_0$ scratch, we choose checkpoints at epoch 1, 2, 3, 4, 5, 10, and 15 and evaluate them on the 4 original tasks from SimplerEnv. Then we pick the best performing checkpoint for each method and roll out the full-scale evaluation in \suite{}. Better performance could be possible if we sweep all checkpoints in \suite{}, but since our evaluation aims to probe the models' generalization capabilities, we believe picking the checkpoint by the original 4 tasks is a more reasonable choice.

\textbf{Threshold For Intention Correctness}
The specific threshold used in the calculation of Intention Correct Rate is set to 5 centimeters. This is empirically determined but is set to be close to other thresholds already exists in the original SimplerEnv.

\section{Prompt PaliGemma for Task Variations}
\label{app:prompt}

Since $\pi_0$ is fine-tuned from PaliGemma, we are interested in knowing whether the PaliGemma VLM itself is able to sort out the task variations before being tuned to action data. We take the ``carrot on plate'', the ``spoon on towel'' and their variation tasks as two sets of example, and the results are displayed in Figure~\ref{fig:paligemma_prompt}.
Specifically, we give the varied task instructions and the first frame of visual observation, and ask PaliGemma to answer in short sentence what the task actually is doing.
As expected, PaliGemma is able to understand many of the task variations, but it also has a strong tendency to repeat the instructions without exactly following the prompts. This causes complication for a targeted measurement as to whether the model does not understand the language variations or it actually understands but tends to literally repeat the previous instruction.
Therefore, we additionally prompt the PaliGemma to specifically answer questions about the variation just like visual question answering. Results on the same ``carrot'' and ``spoon'' examples are in Figure~\ref{fig:paligemma_vqa}.
Results indicate that PaliGemma shows pretty good understanding to the appearance descriptions and commonsense. For example, for the difficult case of ``carrot on plate'' with commonsense and distraction task, the $pi_0$ finetune model make high rate of Wrong Object Attempt as shown in Figure~\ref{fig:distract_carrot}, but the PaliGemma can figure it out, as shown in Figure~\ref{fig:prompt_commons}, providing another piece of evidence that VLA loses VLM's vision-language thinking capability after finetuning.
Of course, PaliGemma makes some mistakes itself, so maybe changing it to a newer and stronger VLM, especially ones with better visual grounding, could mitigate the problem. We leave this investigation as a future work.

\definecolor{light-green}{HTML}{DEF1D3}

\begin{figure}
    \centering
    \begin{subfigure}[h]{\textwidth}
        \centering
        \includegraphics[width=\textwidth]{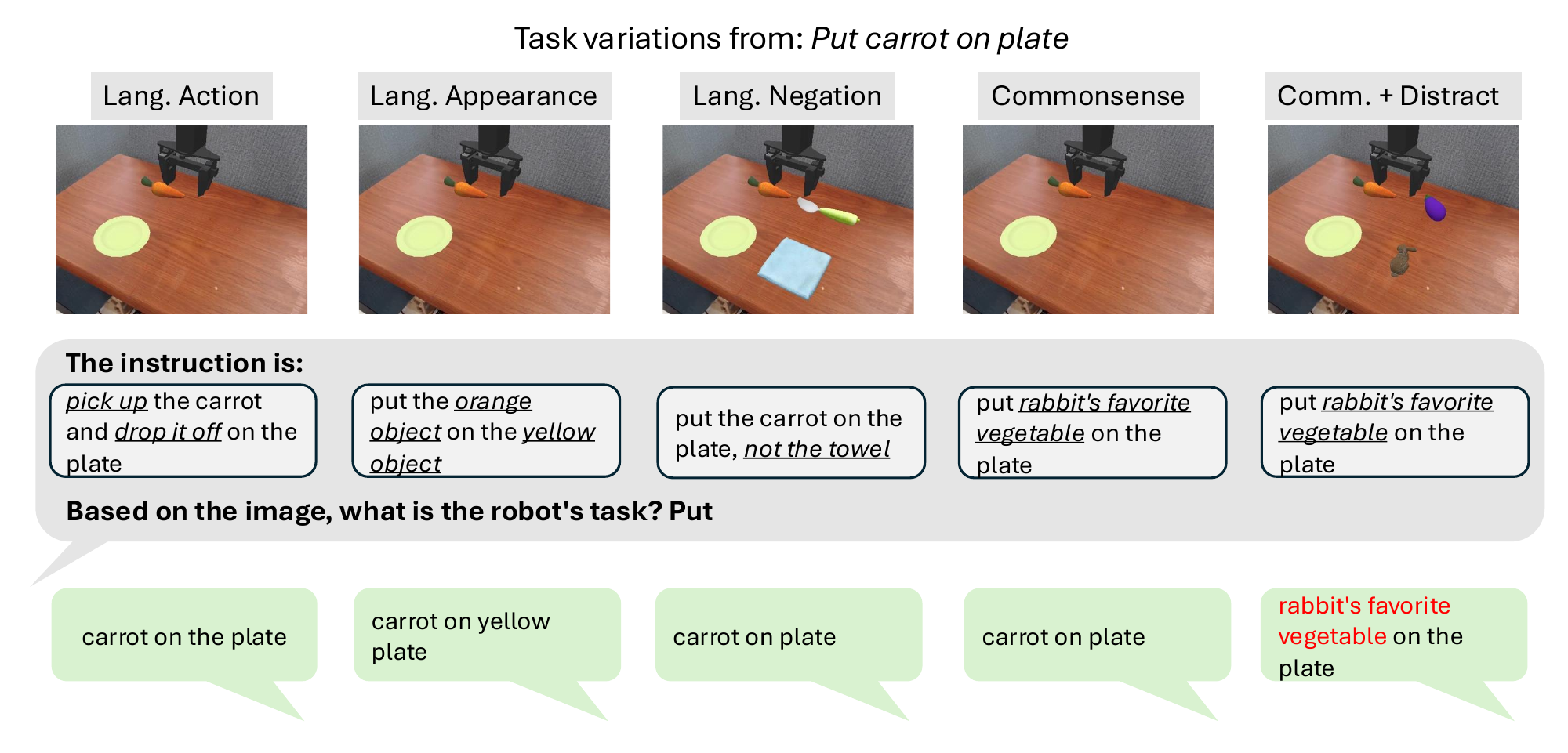}
        \caption{``Carrot on plate'' and its variations.}
        \label{fig:prompt_carrot}
    \end{subfigure}
    \vspace{0.5em}
    \begin{subfigure}[b]{\textwidth}
        \centering
        \includegraphics[width=\textwidth]{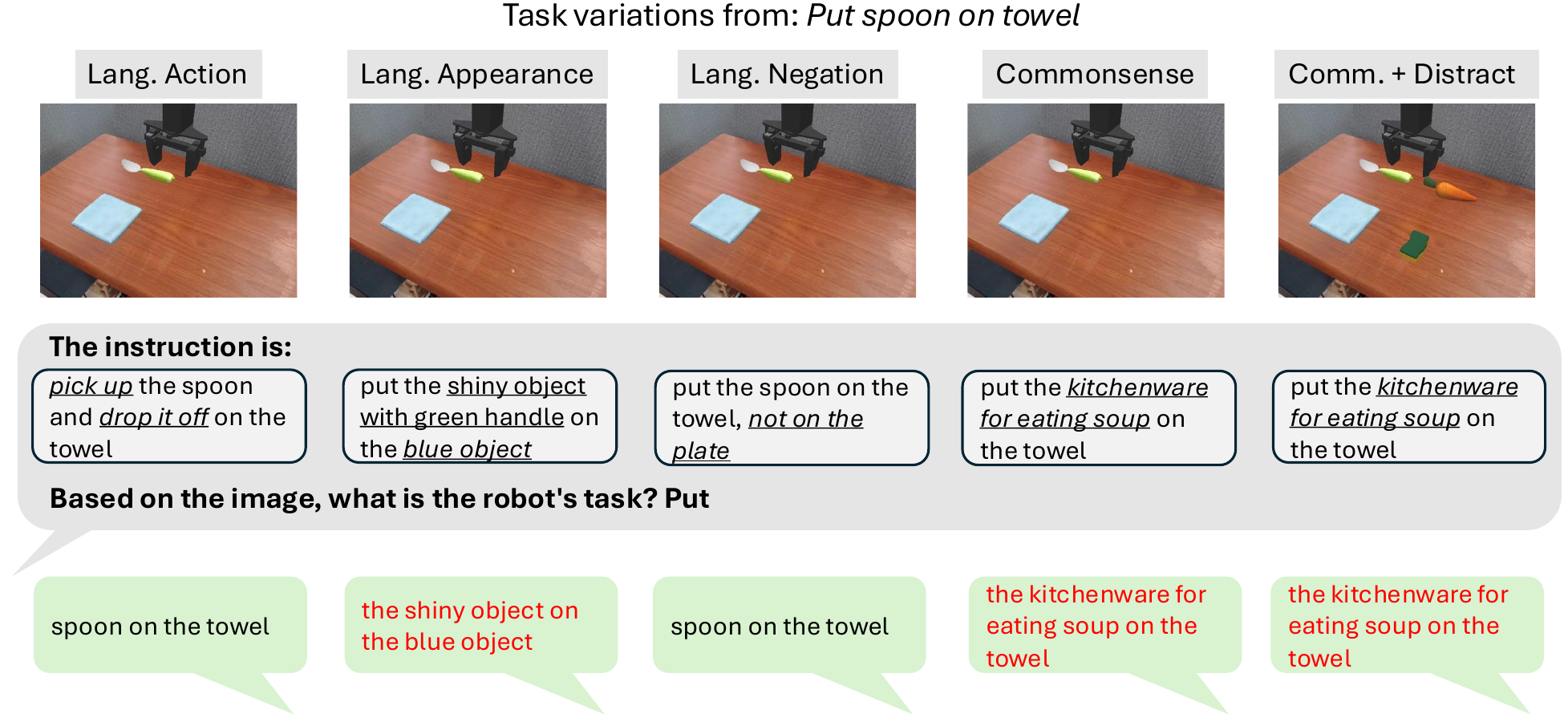}
        \caption{``Spoon on towel'' and its variations.}
        \label{fig:prompt_juice}
    \end{subfigure}
    \caption{Prompting PaliGemma with the task and the first frame of visual observations. Prompts are in \colorbox{oai-gray-300}{gray boxes} where the task variations are \underline{\textit{indicated}}. PaliGemma answers are in \colorbox{light-green}{green boxes} and the wrong answers are in \textcolor{red}{red}.}
    \label{fig:paligemma_prompt}
\end{figure}

\begin{figure}
    \centering
    \begin{subfigure}[h]{\textwidth}
        \centering
        \includegraphics[width=\textwidth]{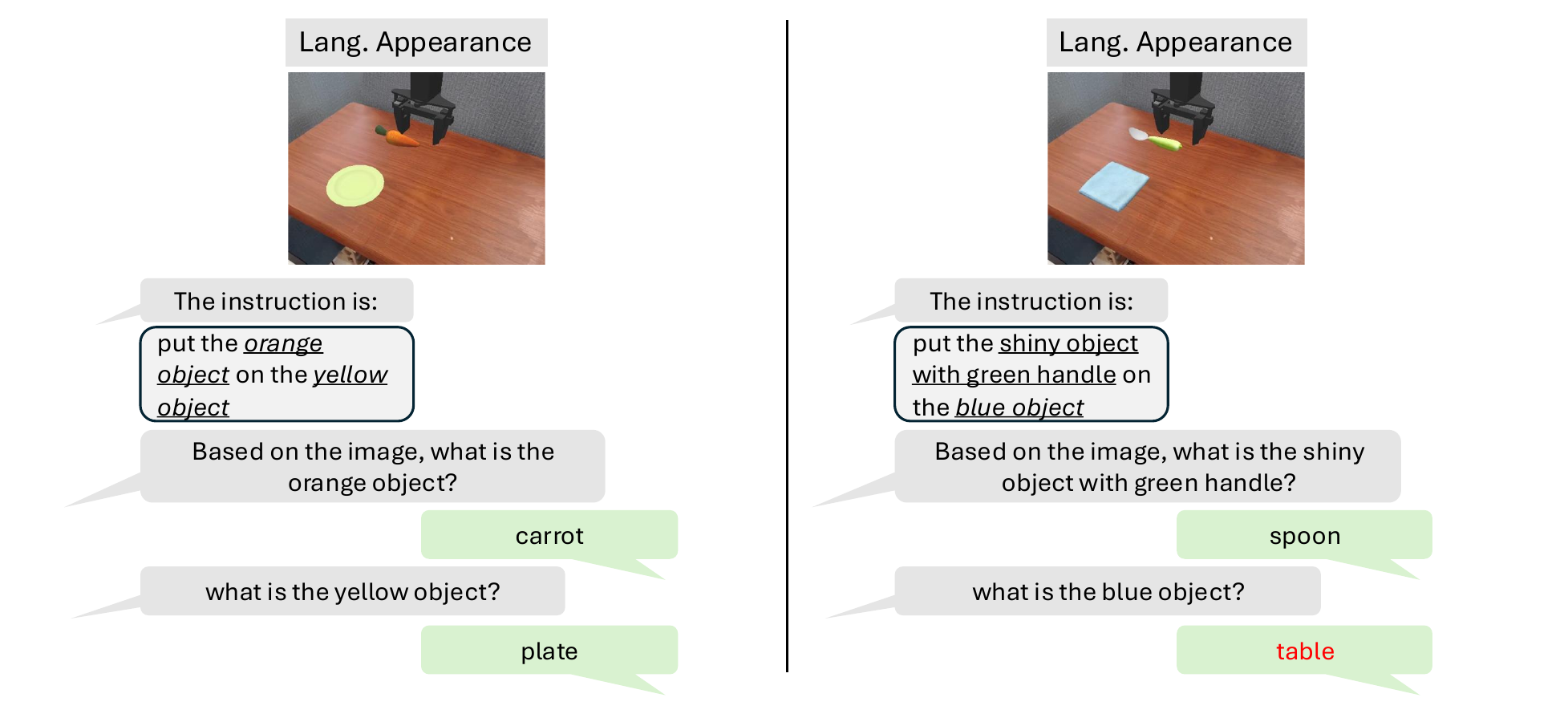}
        \caption{VQA: appearance. Carrot case on the left, spoon case on the right.}
        \label{fig:prompt_appear}
    \end{subfigure}
    \vspace{0.5em}
    \begin{subfigure}[b]{\textwidth}
        \centering
        \includegraphics[width=\textwidth]{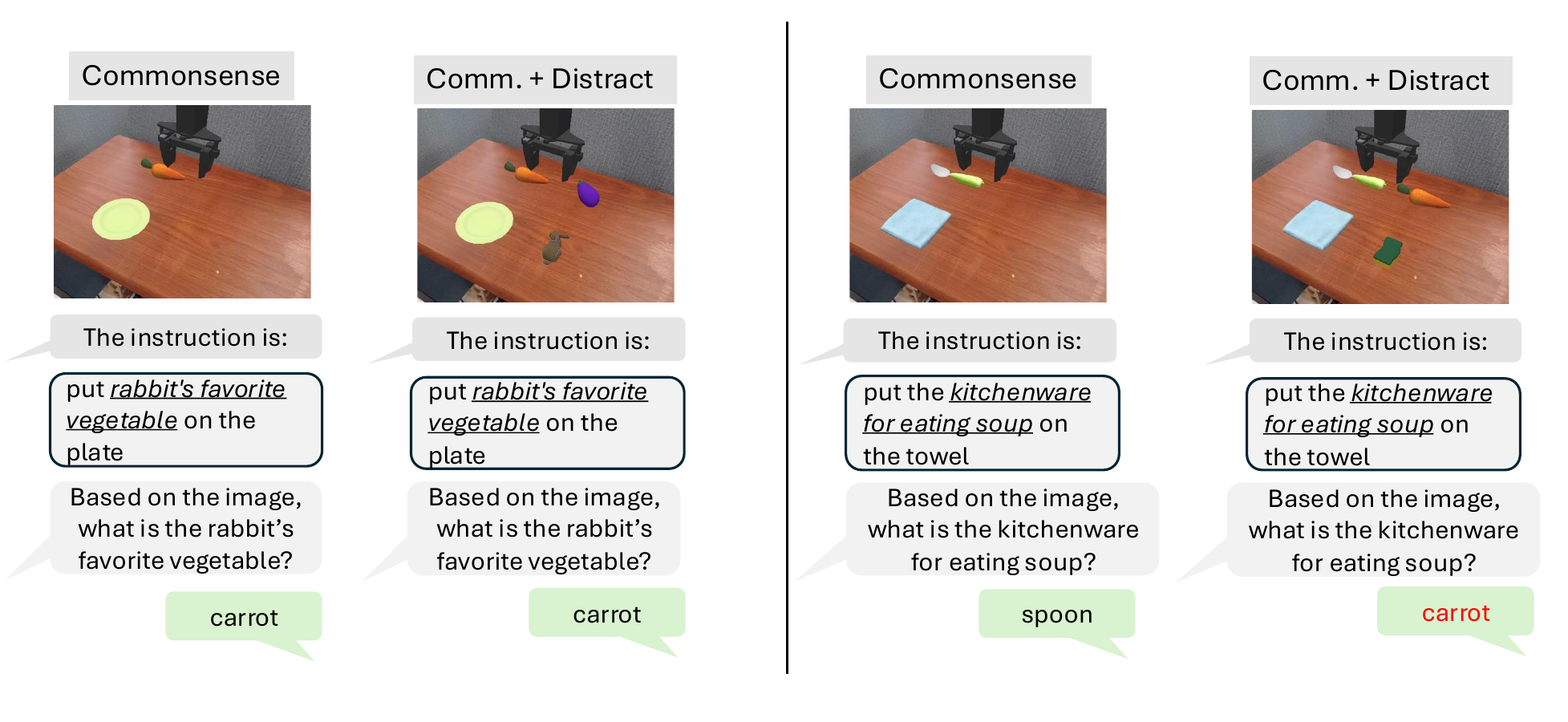}
        \caption{VQA: commonsense. Carrot case on the left, spoon case on the right.}
        \label{fig:prompt_commons}
    \end{subfigure}
    \caption{Prompting PaliGemma to do VQA on the task instructions and the first frame of visual observations. Prompts are in \colorbox{oai-gray-300}{gray boxes} where the task variations are \underline{\textit{indicated}}. PaliGemma answers are in \colorbox{light-green}{green boxes} and the wrong answers are in \textcolor{red}{red}.}
    \label{fig:paligemma_vqa}
\end{figure}
\clearpage
\section{Full Benchmarking Results}
In this part we attach the full results. In addition to the evaluation metrics, we also compute some performance deltas. A map of how the deltas are computed is included in Appendix~\ref{app:delta_performance}.
We will release a plot to better visualize these results in the future.

\label{app:full_results}

\subsection{Original Tasks}
\begin{table}[h]
    \centering
    \caption{Full benchmark results for the original Simpler tasks}
    \resizebox{\textwidth}{!}{%
\begin{tabular}{l|ccc|ccc|ccc|ccc}
    \hline
      Model & Grasp Correct &	Delta &	Delta(\%) &	Intention Correct &	Delta &	Delta(\%) &	Success Rate & Delta & Delta(\%) &	Wrong Obj Attempt &	Delta &	Delta(\%)\\
    \hline
    \multicolumn{13}{c}
    {\textbf{widowx\_carrot\_on\_plate}}\\
    \hline
$\pi_0$ finetune & 0.611 & 0.000 & 0.000 & 1.000 & 0.000 & 0.000 & 0.361 & 0.000 & 0.000 & 0.000 & 0.000 & N/A \\
$\pi_0$ baseline & 0.611 & 0.000 & 0.000 & 1.000 & 0.000 & 0.000 & 0.542 & 0.000 & 0.000 & 0.000 & 0.000 & N/A \\
Spatial VLA  & 0.417 & 0.000 & 0.000 & 1.000 & 0.000 & 0.000 & 0.125 & 0.000 & 0.000 & 0.000 & 0.000 & N/A \\
Magma        & 0.444 & 0.000 & 0.000 & 1.000 & 0.000 & 0.000 & 0.250 & 0.000 & 0.000 & 0.000 & 0.000 & N/A \\
Octo Small   & 0.167 & 0.000 & 0.000 & 0.389 & 0.000 & 0.000 & 0.014 & 0.000 & 0.000 & 0.000 & 0.000 & N/A \\
Octo Base    & 0.222 & 0.000 & 0.000 & 0.750 & 0.000 & 0.000 & 0.014 & 0.000 & 0.000 & 0.000 & 0.000 & N/A \\
\bottomrule
    \end{tabular}}
\end{table}

\subsection{Out-of-Distribution Objects}
\subsubsection{Out-of-Distribution Sources}
\begin{table}[h]
    \centering
    \caption{Full benchmark results for the OOD Source Tasks}
    \resizebox{\textwidth}{!}{%
    \begin{tabular}{l|ccc|ccc|ccc|ccc}
    \hline
      Model & Grasp Correct &	Delta &	Delta(\%) &	Intention Correct &	Delta &	Delta(\%) &	Success Rate & Delta & Delta(\%) &	Wrong Obj Attempt &	Delta &	Delta(\%)\\
    \hline
    \multicolumn{13}{c}{\textbf{widowx\_pepsi\_on\_plate\_clean}}\\
    \hline
$\pi_0$ finetune & 0.639 & 0.028  & 4.549   & 0.875 & -0.125 & -12.497 & 0.444 & 0.083 & 23.087  & 0.000 & 0.000 & N/A \\
$\pi_0$ baseline & 0.875 & 0.264  & 43.190  & 1.000 & 0.000  & 0.000   & 0.722 & 0.181 & 33.335  & 0.000 & 0.000 & N/A \\
Spatial VLA  & 0.875 & 0.458  & 109.983 & 1.000 & 0.000  & 0.000   & 0.333 & 0.208 & 166.640 & 0.000 & 0.000 & N/A \\
Magma        & 0.514 & 0.069  & 15.622  & 0.917 & -0.083 & -8.333  & 0.292 & 0.042 & 16.680  & 0.000 & 0.000 & N/A \\
Octo Small   & 0.236 & 0.069  & 41.680  & 0.417 & 0.028  & 7.140   & 0.014 & 0.000 & 0.000   & 0.000 & 0.000 & N/A \\
Octo Base    & 0.194 & -0.028 & -12.481 & 0.611 & -0.139 & -18.516 & 0.042 & 0.028 & 199.760 & 0.014 & 0.014 & N/A \\
\bottomrule
    \end{tabular}}
\end{table}
\clearpage

\subsubsection{Out-of-Distribution Target}
\begin{table}[h]
    \centering
    \caption{Full benchmark results for the OOD Target tasks}
    \resizebox{\textwidth}{!}{%
    \begin{tabular}{l|ccc|ccc|ccc|ccc}
    \hline
      Model & Grasp Correct &	Delta &	Delta(\%) &	Intention Correct &	Delta &	Delta(\%) &	Success Rate & Delta & Delta(\%) &	Wrong Obj Attempt &	Delta &	Delta(\%)\\
    \hline
    \multicolumn{13}{c}{\textbf{widowx\_carrot\_on\_keyboard\_clean}}\\
    \hline
$\pi_0$ finetune & 0.597 & -0.014 & -2.275  & 1.000 & 0.000  & 0.000   & 0.250 & -0.111 & -30.767  & 0.014 & 0.014 & N/A \\
$\pi_0$ baseline & 0.556 & -0.056 & -9.087  & 1.000 & 0.000  & 0.000   & 0.306 & -0.236 & -43.588  & 0.000 & 0.000 & N/A \\
Spatial VLA  & 0.167 & -0.250 & -59.995 & 0.208 & -0.792 & -79.170 & 0.042 & -0.083 & -66.640  & 0.000 & 0.000 & N/A \\
Magma        & 0.417 & -0.028 & -6.262  & 0.792 & -0.208 & -20.830 & 0.208 & -0.042 & -16.667  & 0.000 & 0.000 & N/A \\
Octo Small   & 0.083 & -0.083 & -50.000 & 0.181 & -0.208 & -53.570 & 0.000 & -0.014 & -100.000 & 0.000 & 0.000 & N/A \\
Octo Base    & 0.167 & -0.056 & -24.992 & 0.583 & -0.167 & -22.219 & 0.000 & -0.014 & -100.000 & 0.000 & 0.000 & N/A \\
\bottomrule
    \end{tabular}}
\end{table}

\subsubsection{Out-of-Distribution Source and Target}
\begin{table}[h]
    \centering
    \caption{Full benchmark results for the OOD Source and Target Tasks}
    \resizebox{\textwidth}{!}{%
    \begin{tabular}{l|ccc|ccc|ccc|ccc}
    \hline
      Model & Grasp Correct &	Delta &	Delta(\%) &	Intention Correct &	Delta &	Delta(\%) &	Success Rate & Delta & Delta(\%) &	Wrong Obj Attempt &	Delta &	Delta(\%)\\
    \hline
    \multicolumn{13}{c}{\textbf{widowx\_coke\_can\_on\_keyboard\_clean}}\\
    \hline
$\pi_0$ finetune & 0.778 & -0.125 & -13.850 & 1.000 & 0.000  & 0.000   & 0.389 & -0.347 & -47.168  & 0.000 & 0.000  & N/A      \\
$\pi_0$ baseline & 0.833 & -0.042 & -4.766  & 1.000 & 0.000  & 0.000   & 0.236 & -0.569 & -70.686  & 0.000 & 0.000  & N/A      \\
Spatial VLA  & 0.167 & -0.625 & -78.944 & 0.167 & -0.833 & -83.330 & 0.083 & -0.375 & -81.824  & 0.000 & 0.000  & N/A      \\
Magma        & 0.431 & -0.264 & -38.000 & 0.972 & 0.028  & 2.944   & 0.139 & -0.222 & -61.534  & 0.000 & -0.014 & -100.000 \\
Octo Small   & 0.111 & -0.181 & -61.909 & 0.167 & -0.389 & -70.001 & 0.000 & -0.056 & -100.000 & 0.000 & 0.000  & N/A      \\
Octo Base    & 0.208 & -0.069 & -24.997 & 0.667 & 0.000  & 0.000   & 0.000 & -0.042 & -100.000 & 0.000 & 0.000  & N/A      \\
\bottomrule
    \end{tabular}}
\end{table}
\clearpage

\subsubsection{Out-of-Distribution Objects Relationship}
\begin{table}[h]
    \centering
    \caption{Full benchmark results for the OOD Relation Tasks}
    \resizebox{\textwidth}{!}{%
    \begin{tabular}{l|ccc|ccc|ccc|ccc}
    \hline
      Model & Grasp Correct &	Delta &	Delta(\%) &	Intention Correct &	Delta &	Delta(\%) &	Success Rate & Delta & Delta(\%) &	Wrong Obj Attempt &	Delta &	Delta(\%)\\
    \hline
    \multicolumn{13}{c}{\textbf{widowx\_carrot\_on\_sponge\_clean}}\\
    \hline
$\pi_0$ finetune & 0.65 & 0.04  & 6.82   & 1.00 & 0.00  & 0.00   & 0.28 & -0.08 & -23.08  & 0.01 & 0.01 & N/A \\
$\pi_0$ baseline & 0.67 & 0.06  & 9.09   & 1.00 & 0.00  & 0.00   & 0.60 & 0.06  & 10.25   & 0.01 & 0.01 & N/A \\
Spatial VLA  & 0.38 & -0.04 & -10.01 & 0.96 & -0.04 & -4.17  & 0.17 & 0.04  & 33.36   & 0.00 & 0.00 & N/A \\
Magma        & 0.17 & -0.28 & -62.50 & 0.57 & -0.43 & -43.06 & 0.11 & -0.14 & -55.55  & 0.03 & 0.03 & N/A \\
Octo Small   & 0.06 & -0.11 & -66.66 & 0.15 & -0.24 & -60.71 & 0.00 & -0.01 & -100.00 & 0.00 & 0.00 & N/A \\
Octo Base    & 0.19 & -0.03 & -12.50 & 0.65 & -0.10 & -12.96 & 0.00 & -0.01 & -100.00 & 0.01 & 0.01 & N/A \\
\bottomrule
    \end{tabular}}
\end{table}

\subsection{Language Complexity}
\subsubsection{Language Action}
\begin{table}[h]
    \centering
    \caption{Full benchmark results for the Language Action tasks}
    \resizebox{\textwidth}{!}{%
    \begin{tabular}{l|ccc|ccc|ccc|ccc}
    \hline
      Model & Grasp Correct &	Delta &	Delta(\%) &	Intention Correct &	Delta &	Delta(\%) &	Success Rate & Delta & Delta(\%) &	Wrong Obj Attempt &	Delta &	Delta(\%)\\
    \hline
    \multicolumn{13}{c}{\textbf{widowx\_carrot\_on\_plate\_lang\_action}}\\
    \hline
$\pi_0$ finetune & 0.36 & -0.25 & -40.91  & 0.81 & -0.19 & -19.44 & 0.11 & -0.25 & -69.23  & 0.00 & 0.00 & N/A \\
$\pi_0$ baseline & 0.60 & -0.01 & -2.27   & 1.00 & 0.00  & 0.00   & 0.54 & 0.00  & 0.00    & 0.00 & 0.00 & N/A \\
Spatial VLA  & 0.00 & -0.42 & -100.00 & 0.04 & -0.96 & -95.83 & 0.00 & -0.13 & -100.00 & 0.00 & 0.00 & N/A \\
Magma        & 0.31 & -0.14 & -31.25  & 0.89 & -0.11 & -11.11 & 0.14 & -0.11 & -44.45  & 0.03 & 0.03 & N/A \\
Octo Small   & 0.17 & 0.00  & 0.00    & 0.46 & 0.07  & 17.86  & 0.00 & -0.01 & -100.00 & 0.01 & 0.01 & N/A \\
Octo Base    & 0.03 & -0.19 & -87.50  & 0.35 & -0.40 & -53.70 & 0.00 & -0.01 & -100.00 & 0.01 & 0.01 & N/A \\
\bottomrule
    \end{tabular}}
\end{table}
\clearpage

\subsubsection{Language Negation}
\begin{table}[h]
    \centering
    \caption{Full benchmark results for the Language Negation tasks}
    \resizebox{\textwidth}{!}{%
    \begin{tabular}{l|ccc|ccc|ccc|ccc}
    \hline
      Model & Grasp Correct &	Delta &	Delta(\%) &	Intention Correct &	Delta &	Delta(\%) &	Success Rate & Delta & Delta(\%) &	Wrong Obj Attempt &	Delta &	Delta(\%)\\
    \hline
    \multicolumn{13}{c}{\textbf{widowx\_carrot\_on\_plate\_lang\_neg}}\\
    \hline
$\pi_0$ finetune & 0.54 & -0.07 & -11.37 & 1.00 & 0.00  & 0.00   & 0.11 & -0.25 & -69.22 & 0.00 & 0.00 & N/A \\
$\pi_0$ baseline & 0.72 & 0.11  & 18.19  & 1.00 & 0.00  & 0.00   & 0.51 & -0.03 & -5.13  & 0.00 & 0.00 & N/A \\
Spatial VLA  & 0.42 & 0.00  & 0.00   & 0.92 & -0.08 & -8.33  & 0.38 & 0.25  & 200.00 & 0.00 & 0.00 & N/A \\
Magma        & 0.37 & -0.07 & -15.64 & 0.83 & -0.17 & -16.67 & 0.07 & -0.18 & -72.21 & 0.00 & 0.00 & N/A \\
Octo Small   & 0.17 & 0.00  & 0.00   & 0.54 & 0.15  & 39.28  & 0.01 & 0.00  & 0.00   & 0.06 & 0.06 & N/A \\
Octo Base    & 0.18 & -0.04 & -18.74 & 0.65 & -0.10 & -12.96 & 0.01 & 0.00  & 0.00   & 0.01 & 0.01 & N/A \\
\bottomrule
    \end{tabular}}
\end{table}

\subsubsection{Language Appearance}
\begin{table}[h]
    \centering
    \caption{Full benchmark results for the Language Appearance tasks}
    \resizebox{\textwidth}{!}{%
    \begin{tabular}{l|ccc|ccc|ccc|ccc}
    \hline
      Model & Grasp Correct &	Delta &	Delta(\%) &	Intention Correct &	Delta &	Delta(\%) &	Success Rate & Delta & Delta(\%) &	Wrong Obj Attempt &	Delta &	Delta(\%)\\
    \hline
    \multicolumn{13}{c}{\textbf{widowx\_carrot\_on\_plate\_lang\_color											}}\\
    \hline
$\pi_0$ finetune & 0.36 & -0.25 & -40.91 & 0.99 & -0.01 & -1.39  & 0.28 & -0.08 & -23.08  & 0.01 & 0.01 & N/A \\
$\pi_0$ baseline & 0.58 & -0.03 & -4.54  & 1.00 & 0.00  & 0.00   & 0.56 & 0.01  & 2.57    & 0.00 & 0.00 & N/A \\
Spatial VLA  & 0.25 & -0.17 & -40.00 & 0.42 & -0.58 & -58.33 & 0.04 & -0.08 & -66.64  & 0.00 & 0.00 & N/A \\
Magma        & 0.25 & -0.19 & -43.75 & 0.88 & -0.13 & -12.50 & 0.08 & -0.17 & -66.68  & 0.00 & 0.00 & N/A \\
Octo Small   & 0.01 & -0.15 & -91.66 & 0.14 & -0.25 & -64.28 & 0.00 & -0.01 & -100.00 & 0.00 & 0.00 & N/A \\
Octo Base    & 0.04 & -0.18 & -81.25 & 0.35 & -0.40 & -53.70 & 0.00 & -0.01 & -100.00 & 0.04 & 0.04 & N/A \\
\bottomrule
    \end{tabular}}
\end{table}

\clearpage

\subsection{Vision-Language Thinking}
\subsubsection{Object Distraction}
\begin{table}[h]
    \centering
    \caption{Full benchmark results for the Object Distraction tasks}
    \resizebox{\textwidth}{!}{%
    \begin{tabular}{l|ccc|ccc|ccc|ccc}
    \hline
      Model & Grasp Correct &	Delta &	Delta(\%) &	Intention Correct &	Delta &	Delta(\%) &	Success Rate & Delta & Delta(\%) &	Wrong Obj Attempt &	Delta &	Delta(\%)\\
    \hline
    \multicolumn{13}{c}{\textbf{widowx\_carrot\_on\_plate\_distract											}}\\
    \hline
$\pi_0$ finetune & 0.708 & 0.097  & 15.911  & 1.000 & 0.000  & 0.000  & 0.222 & -0.139 & -38.457 & 0.000 & 0.000 & N/A \\
$\pi_0$ baseline & 0.639 & 0.028  & 4.549   & 1.000 & 0.000  & 0.000  & 0.542 & 0.000  & 0.000   & 0.028 & 0.028 & N/A \\
Spatial VLA  & 0.458 & 0.042  & 9.983   & 0.958 & -0.042 & -4.170 & 0.208 & 0.083  & 66.640  & 0.000 & 0.000 & N/A \\
Magma        & 0.583 & 0.139  & 31.236  & 0.944 & -0.056 & -5.557 & 0.236 & -0.014 & -5.547  & 0.000 & 0.000 & N/A \\
Octo Small   & 0.153 & -0.014 & -8.340  & 0.583 & 0.194  & 49.996 & 0.014 & 0.000  & 0.000   & 0.014 & 0.014 & N/A \\
Octo Base    & 0.181 & -0.042 & -18.737 & 0.736 & -0.014 & -1.845 & 0.014 & 0.000  & 0.000   & 0.014 & 0.014 & N/A \\
\bottomrule
    \end{tabular}}
\end{table}
\clearpage

\subsubsection{Commonsense}
\begin{table}[h]
    \centering
    \caption{Full benchmark results for the Commonsense tasks}
    \resizebox{\textwidth}{!}{%
    \begin{tabular}{l|ccc|ccc|ccc|ccc}
    \hline
      Model & Grasp Correct &	Delta &	Delta(\%) &	Intention Correct &	Delta &	Delta(\%) &	Success Rate & Delta & Delta(\%) &	Wrong Obj Attempt &	Delta &	Delta(\%)\\
    \hline
    \multicolumn{13}{c}{\textbf{widowx\_carrot\_on\_plate\_lang\_common}}\\
    \hline
$\pi_0$ finetune & 0.153 & -0.458 & -74.996  & 0.875 & -0.125 & -12.500 & 0.083 & -0.278 & -76.922  & 0.028 & 0.028 & N/A \\
$\pi_0$ baseline & 0.431 & -0.181 & -29.542  & 0.986 & -0.014 & -1.390  & 0.292 & -0.250 & -46.148  & 0.000 & 0.000 & N/A \\
Spatial VLA  & 0.167 & -0.250 & -59.995  & 1.000 & 0.000  & 0.000   & 0.042 & -0.083 & -66.640  & 0.000 & 0.000 & N/A \\
Magma        & 0.611 & 0.167  & 37.491   & 1.000 & 0.000  & 0.000   & 0.278 & 0.028  & 11.107   & 0.000 & 0.000 & N/A \\
Octo Small   & 0.028 & -0.139 & -83.340  & 0.111 & -0.278 & -71.432 & 0.000 & -0.014 & -100.000 & 0.000 & 0.000 & N/A \\
Octo Base    & 0.000 & -0.222 & -100.000 & 0.139 & -0.611 & -81.479 & 0.000 & -0.014 & -100.000 & 0.000 & 0.000 & N/A \\
\bottomrule
    \end{tabular}}
\end{table}
\clearpage

\subsubsection{Commonsense and Distraction}
\begin{table}[h]
    \centering
    \caption{Full benchmark results for the Commonsense and Distraction tasks}
    \resizebox{\textwidth}{!}{%
    \begin{tabular}{l|ccc|ccc|ccc|ccc}
    \hline
      Model & Grasp Correct &	Delta &	Delta(\%) &	Intention Correct &	Delta &	Delta(\%) &	Success Rate & Delta & Delta(\%) &	Wrong Obj Attempt &	Delta &	Delta(\%)\\
    \hline
    \multicolumn{13}{c}{\textbf{widowx\_spoon\_on\_towel\_lang\_common\_distract}}\\
    \hline
$\pi_0$ finetune & 0.208 & 0.069  & 49.988  & 0.514 & 0.000  & -0.006  & 0.069 & 0.000  & 0.000    & 0.042 & 0.042 & N/A     \\
$\pi_0$ baseline & 0.569 & -0.111 & -16.330 & 1.000 & 0.000  & 0.000   & 0.167 & -0.389 & -70.007  & 0.014 & 0.000 & 0.000   \\
Spatial VLA  & 0.250 & -0.167 & -40.005 & 0.625 & -0.333 & -34.780 & 0.042 & -0.208 & -83.320  & 0.250 & 0.250 & N/A     \\
Magma        & 0.222 & -0.069 & -23.797 & 0.889 & -0.042 & -4.478  & 0.069 & -0.042 & -37.474  & 0.014 & 0.014 & N/A     \\
Octo Small   & 0.042 & -0.014 & -25.015 & 0.319 & -0.014 & -4.160  & 0.000 & -0.014 & -100.000 & 0.014 & 0.014 & N/A     \\
Octo Base    & 0.042 & 0.000  & 0.000   & 0.278 & 0.083  & 42.852  & 0.000 & 0.000  & N/A      & 0.111 & 0.097 & 699.281 \\
\bottomrule
    \end{tabular}}
\end{table}

\clearpage
\section{How to Calculate Delta}
\label{app:delta_performance}
The deltas in all the metrics are calculated by subtracting the performance in the tasks listed in the left column of Table \ref{tab:full-task-mapping} from the performance in the corresponding task listed in the right column.
\begin{table}[ht]
\centering
\caption{Full mapping from variant task names to canonical task names.}
\resizebox{\textwidth}{!}{%
\begin{tabular}{|l|l|}
\hline
\textbf{Variant Task Name} & \textbf{Canonical Task Name} \\
\hline
widowx\_spoon\_on\_towel & widowx\_spoon\_on\_towel \\
widowx\_carrot\_on\_plate & widowx\_carrot\_on\_plate \\
widowx\_stack\_cube & widowx\_stack\_cube \\
widowx\_put\_eggplant\_in\_basket & widowx\_put\_eggplant\_in\_basket \\
widowx\_cube\_on\_plate\_clean & widowx\_carrot\_on\_plate \\
widowx\_small\_plate\_on\_green\_cube\_clean & widowx\_cube\_on\_plate\_clean \\
widowx\_coke\_can\_on\_plate\_clean & widowx\_carrot\_on\_plate \\
\hline
\end{tabular}
}

\label{tab:full-task-mapping}
\end{table}